\documentclass[10pt,journal,compsoc]{IEEEtran}

\ifCLASSOPTIONcompsoc
  \usepackage[nocompress]{cite}
\else
  \usepackage{cite}
\fi
\ifCLASSINFOpdf
\else
\fi

\usepackage{algorithmic}

\usepackage{times}
\usepackage{soul}
\usepackage{url}
\usepackage[hidelinks]{hyperref}
\usepackage[utf8]{inputenc}
\usepackage[small]{caption}
\usepackage{graphicx}
\usepackage{amsmath}

\usepackage{amsthm}
\usepackage{hhline}
\usepackage{multirow}
\usepackage{booktabs}
\usepackage[ruled,vlined]{algorithm2e}
\usepackage{algorithmic}
\usepackage{adjustbox}

\usepackage{mathtools}
\usepackage{amssymb}
\usepackage{subfig}
\usepackage{color}

\newcommand{\std}[1]{\scriptscriptstyle\pm{#1}}
\DeclareMathOperator{\Tr}{tr}

\DeclarePairedDelimiter{\norm}{\lVert}{\rVert}

\begin{document}
%
\title{3D Skeleton-based Human Motion Prediction with Manifold-Aware GAN}
%
%
%
%
\author{
Baptiste~Chopin,
Naima~Otberdout,
Mohamed~Daoudi,~\IEEEmembership{Senior,~IEEE,}
Angela Bartolo
\IEEEcompsocitemizethanks{ \IEEEcompsocthanksitem B. Chopin and N. Otberdout are with Univ. Lille, CNRS, Inria, Centrale Lille, UMR 9189 CRIStAL, F-59000 Lille, France. E-mail:. \{baptiste.chopin,naima.otberdout\}@univ-lille.fr
\IEEEcompsocthanksitem M. Daoudi is with IMT Nord Europe, Institut Mines-Télécom, Univ. Lille, Centre for Digital Systems, F-59000 Lille, France, and Univ. Lille, CNRS, Centrale Lille, Institut Mines-Télécom, UMR 9189 CRIStAL, F-59000 Lille, France, E-mail: mohamed.daoudi@imt-nord-europe.fr
\IEEEcompsocthanksitem Angela Bartolo is with Univ. Lille, CNRS, UMR 9193 SCALab, F-59000 Lille, France. E-mail: angela.bartolo@univ-lille.fr}}%

\markboth{Journal of \LaTeX\ Class Files,~Vol.~14, No.~8, August~2015}%
{Shell \MakeLowercase{\textit{et al.}}: Bare Advanced Demo of IEEEtran.cls for IEEE Biometrics Council Journals}
%



\IEEEtitleabstractindextext{%
\begin{abstract}


In this work we propose a novel solution for 3D skeleton-based human motion prediction. The objective of this task consists in forecasting future human poses based on a prior skeleton pose sequence. This involves solving two main challenges still present in recent literature; (1) discontinuity of the predicted motion which results in unrealistic motions and (2) performance deterioration in long-term horizons resulting from error accumulation across time. We tackle these issues by using a compact manifold-valued representation of 3D human skeleton motion. Specifically, we model the temporal evolution of the 3D poses as trajectory, what allows us to map human motions to single points on a sphere manifold. Using such a compact representation avoids error accumulation and provides robust representation for long-term prediction while ensuring  the  smoothness  and  the  coherence  of  the  whole  motion. To learn these non-Euclidean representations, we build a manifold-aware Wasserstein generative adversarial model that captures the temporal and spatial dependencies of human motion through different losses. Experiments have been conducted on CMU MoCap and Human 3.6M datasets and demonstrate the superiority of our approach over the state-of-the-art both in short and long term horizons. The smoothness of the generated motion is highlighted in the qualitative results.

\end{abstract}

}

\maketitle

\IEEEdisplaynontitleabstractindextext

%
\IEEEpeerreviewmaketitle

\ifCLASSOPTIONcompsoc
\IEEEraisesectionheading{\section{Introduction}\label{sec:introduction}}
\else
\section{Introduction}
\label{sec:introduction}
\fi

\begin{figure*}[!ht]
    \centering
    \includegraphics[width=0.9\linewidth,height=8.5cm]{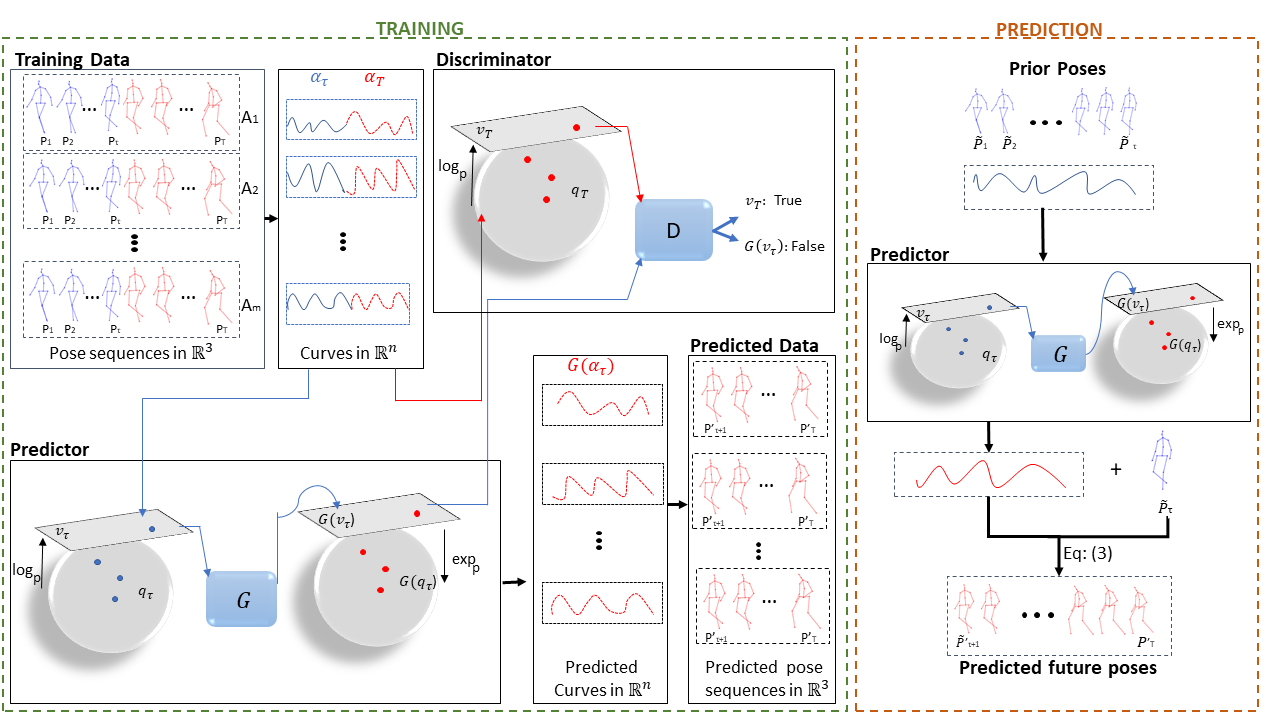}
    \caption{Overview of the human motion training and prediction processes. Given a pose sequence history represented as a curve, then mapped to a single point in a hypersphere. The predictor maps the input point to a tangent space, then feeds it to the network $\mathcal{G}$ that predicts the future motion as a vector in $T_{\mu}(\mathcal{C})$. During training a discriminator is used to compare the mapped points from the ground truth to the generated ones. Exponential operator maps this vector to $\mathcal{C}$, before transforming it to a curve representing a motion. The predicted motion is transformed into a 3D human pose sequence corresponding to the future poses of the prior ones.}
    \label{fig:overview}
\end{figure*}

%
%
%
%
\IEEEPARstart{T}{he} problem of forecasting future human motion play a vital role in many applications in computer vision and robotics, such as human-robot interaction~\cite{KoppulaIEEEROS2013}, autonomous driving~\cite{DBLP:journals/tiv/PadenCYYF16}  and computer graphics~\cite{kovar2008motion}. In this work, we propose a 
predictive model for short and long-term future 3D skeleton poses given an initial prior history.
Addressing this issue involves two main challenges: How to represent the temporal evolution of the human motion to ensure the smoothness of the predicted sequences? and how to take the spatial correlations between human joints into account to avoid implausible poses?  \\
\indent
Because of the explosion of deep learning and the availability of large scale datasets for human motion analysis, deep learning models have been widely exploited to address the problem of human motion prediction and especially Recurrent Neural Networks (RNN)~\cite{DBLP:conf/iccv/FragkiadakiLFM15,jain_structural-rnn_2016, ghosh2017learning,  martinez_human_2017}. Indeed, RNN-based approaches achieved good advance in term of accuracy, however, the motions predicted with these methods present significant discontinuities due to the frame-by-frame regression process that  discourage the global smoothness of the motion. In addition, recurrent models suffer from error accumulation across time, which
increase error and worsen long-term forecasting performance.
To remedy this, more recent works avoid these models and explore feed-forward networks instead. Including CNN~\cite{li_convolutional_2018}, GNN~\cite{mao2020learning}  and fully-connected networks~\cite{butepage2017deep}. Thanks to their hierarchical structure, feed-forward networks can better deal with the spatial correlations of human joints than RNNs. However, an additional strategy is required to encode the temporal information when using these models. To face this issue, an interesting idea was to model the human motion as trajectory~\cite{MaoICCV19},~\cite{BerrettiACMTOM}.\\
In this work, we follow the idea of modeling motions as trajectories in time but in a different context from the previous work. Among the benefits of our representation, the possibility to map these trajectories to single compact points on a manifold, which helps preserving the continuity and the smoothness of the predicted motions. Besides, the compact representation avoids the problem of the error accumulation across time and makes our approach suitable for long-term prediction as illustrated in Figure~\ref{fig:introduction}. Nevertheless, the challenge here is that the resulting representations are manifold-valued data that cannot be manipulated with traditional generative models in a straightforward manner. 
To face this challenge, we introduce in this paper, a manifold-aware Wasserstein Generative Adversarial Networks (WGAN) that predict future skeleton poses given the input prior motion sequence that is encoded as a manifold-valued data. 
The spatial dependencies between human joints are taken into consideration in our method through additional loss functions that add more constraints on the predicted skeleton poses to ensure their plausibility. An overview of our prediction process is illustrated in Figure~\ref{fig:overview}.\\
The contribution of this work can be summarized as follows: (1) To the best of our knowledge, we are the first to propose an approach that exploits compact manifold-valued representation for human motion prediction. By doing so, we model both the temporal and the spatial dependencies involved in human motion, resulting in smooth motions and plausible poses in long-term horizons. (2) We propose a predictive manifold-aware WGAN for motion prediction. (3) We propose a new loss function based on Gram matrix of the 3D poses that avoids predicting implausible poses. (4) Experimental results on Human 3.6M and the CMU MoCap datasets show quantitatively and visually the effectiveness of our method for short-term and long-term prediction.\\

We presented some preliminary ideas of this work in~\cite{chopin2021human}.
With respect to~\cite{chopin2021human}, this paper provides more theoretical details about the proposed approach, it includes also new figures and more discussion. Furthermore, we present in this paper more results that further demonstrate the superiority of our solution over the state-of-the-art. While in~\cite{chopin2021human}, we compared the approaches based on joints position, we present here a new metric to evaluate the smoothness and the temporal evolution of the predicted motion. We provide a new qualitative evaluation for our ablation study to highlight the importance of the different losses of our method.  We also demonstrate in this paper the ability of our method to predict longer sequences by recursive generation.   \\


\section{Related Work}\label{sec:RelatedWork}
\textbf{Human Motion Prediction with Deep Learning}. Given that  the task of human motion prediction is a temporal dependent problem, recurrent models (RNN) were the first potential solution to be investigated, hence several works applied RNN and their variants to tackle this task. In~\cite{DBLP:conf/iccv/FragkiadakiLFM15}, the authors proposed a model that incorporates a nonlinear encoder and decoder before and after recurrent layers. Their approach suffers from error accumulation and discontinuity between the last frame of the prior and the first frames of the generated sequence. Moreover, their approach only capture the temporal dependencies but ignore the spatial correlations between articulations. To deal with this problem, \cite{jain_structural-rnn_2016} proposed a Structural-RNN model relying on high-level spatio-temporal graphs. \cite{ghosh2017learning} take a different direction to minimize the error accumulation effect in RNNs; they  
used a feed forward network for pose filtering and a RNN for temporal filtering. However, this strategy only minimizes the accumulated error that still exists and deteriorates the performance of recurrent models in long-term prediction.
Alternatively, more recent works exploit feed-forward networks. To model the temporal evolution with these models, various strategies have been suggested. In ~\cite{li_convolutional_2018,butepage2017deep}, convolution across time was exploited to model the temporal dependencies with convolution networks, while~\cite{MaoICCV19} adopt Discrete Cosine Transform to encode the motion as trajectory. Graph neural networks were also applied for motion prediction~\cite{mao2020learning, jain_structural-rnn_2016} as a suitable tool to model the spatial correlations involved between the articulations.\\ In this paper, we take a completely different direction and we propose to deal with human motion by exploiting a manifold-valued representation with generative adversarial models. \\

\indent\textbf{Generative Adversarial Networks (GANs)}: GANs have been also exploited to address the problem of human motion prediction in~\cite{ferrari_adversarial_2018} and~\cite{barsoum2018hp}, however, in order to model the temporal dependencies involved, they build their generator on RNN structures. In this way, the error accumulation problem is present in their model which may deteriorate its performance in the long-term. In our work we completely discard recurrent models by adopting a compact representation of the human motion. \\ 
Motivated by the interest of manifold-valued images in a variety of applications, \cite{ZhiwuHuang2017} proposed manifold-aware WGAN. Inspired from this work, we build a manifold-aware WGAN that predict the future points of a poses trajectory given previous pose sequence. However, our model is different from the one proposed in~\cite{ZhiwuHuang2017} in two ways. Firstly, instead of unsupervised image generation from a vector noise, our model addresses the problem of predicting future manifold-valued representations from a manifold-valued inputs. In addition, we propose different objective functions to train our model on the task at hand.\\

\indent
\textbf{Modeling Human Motions as Trajectories on a Riemannian Manifold}: While our present work is the first that explores the benefit of manifold-valued trajectories  for human motion prediction, representing 3D human poses and their temporal evolution as trajectories on a manifold was adopted in many recent works for action recognition. 
Different manifolds were considered in different studies ~\cite{Turagacvpr2009}, ~\cite{Boulbaba2016PAMI}, ~\cite{KacemPAMI2020}. 
 More related to our work, in~\cite{devanne20143}, a human action is interpreted as a parametrized curve and is seen as a single point on the sphere by computing its Square Root Velocity Function (SRVF). Accordingly, different actions were classified based on the distance between their associated points on the sphere. All papers mentioned above show the effectiveness of  motion modeling as a trajectory in action recognition. Motivated by this fact, we show in this paper the interest of using such representation to address the recent challenges that still encountered in human motion prediction.
 
\section{Human Motion Modeling}\label{sec:motionModeling}
Two 3D skeleton representations were adopted for human motion prediction; angles based and 3D coordinates based representations. The first one models each joint by its rotation in term of Euler angles, while  the second representation uses the 3D coordinates of the joints. More recently,~\cite{mao2020learning}, showed in their experiments that the angles based representation where two different sets of angles can represent the exact same pose, leads to ambiguous results and cannot provide a fair and reliable comparison. Motivated by this, we use 3D joint coordinates to represent our skeleton poses.

\subsection{Representation of Pose Sequences as Trajectories in $\mathbb{R}^n$ }
Let $k$ be the number of joints that compose the skeleton, we represent $P_t$ the pose of the skeleton at frame $t$ by a n-dimensional tuple: $P_t=[x_1(t), y_1(t), z_1(t) \dots x_{k}(t), y_{k}(t), z_{k}(t)]^T ,$
\noindent The pose $P_t$ encodes the positions of $k$ distinct joints in $3$ dimensions. Consequently, an action sequence of length $T$ frames, can be described as a sequence $\{P_1, P_2 \dots ,P_T\}$, where $P_i  \in \mathbb{R}^n$ and $n=3 \times k$.

\noindent This sequence represents the evolution of the action over time and can be considered as a result of sampling a continuous curve in $\mathbb{R}^n$. Based on this consideration, we model in what follows, each pose sequence of a skeleton, as a continuous curve in $\mathbb{R}^n$ that describes the continuous evolution of the sequence over time. \\
Let us represent the curve describing a pose sequence by a continuous parameterized function $ \alpha (t): I =[0, 1] \rightarrow \mathbb{R}^n$. In this work, we formulate the problem of human motion prediction given the first consecutive frames of the action as the problem of predicting the possible next points of the curve describing these first frames. More formally, the problem of predicting the future poses $\left\{P_{\tau+1},P_{\tau+2}, \dots ,P_T \right\}$, given the first $\tau$ consecutive skeleton poses $\left\{P_1, P_2, \dots ,P_\tau \right\}$, where $\tau < T$, is formulated as the problem of predicting $ \alpha (t)_{t=\tau+1 \dots T}$ given $ \alpha (t)_{t=1 \dots \tau}$, such that, $ \alpha (t)$ is the continuous function representing the curve associated to the pose sequence $\left\{P_1,P_2, \dots ,P_T \right\}$.

\subsection{Representation of Human Motions as Elements in a Hypersphere $\mathcal{C}$ }
For the purpose of modeling and studying our curves, we adopt square-root velocity function (SRVF) proposed in~\cite{SrivastavaKJJ11}. It was successfully exploited for human action recognition~\cite{devanne20143}, 3D face recognition~\cite{drira20133d} and facial expression generation~\cite{OtberdoutPAMI2020}. Conveniently for us, this function maps each curve $\alpha(t)$ to one point in a hypersphere which provides a compact representation of the human motion. Specifically, for a given curve $ \alpha (t): I  \rightarrow \mathbb{R}^n$, the  square-root velocity function (SRVF) $ q(t): I\rightarrow \mathbb{R}^n$ is defined by the formula

\begin{equation}
q(t)= \begin{cases}\frac{\dot{\alpha}(t)}{\sqrt{\|\dot{\alpha}(t)\|_{}}}, & \text { if }\|\dot{\alpha}(t)\| \neq 0 \\ 0, & \text { if }\|\dot{\alpha}(t)\|=0\end{cases}
\end{equation}


\noindent where, $\|\cdot\|$ is the Euclidean 2-norm in $\mathbb{R}^n$. 
We can easily recover the curve (\emph{i.e}, pose sequence) $\alpha(t)$ from the generated SRVF (\emph{i.e}, dynamic information) $q(t)$ by,
\begin{equation}
\label{eq:curve_eq}
\alpha(t) = \int_0^t \norm{q(s)}q(s)ds + \alpha(0)\; ,
\end{equation}
\noindent where $\alpha(0)$ is the skeleton pose at the initial time step which corresponds in our case to the final time step of the history. 
In order to remove the scale variability of the curves, we scale them to be of length $1$. Consequently, the SRVF corresponding to these curves are elements of a unit hypersphere in the Hilbert manifold $\mathbb{L}^2(I,\mathbb{R}^{n})$ as explained in~\cite{SrivastavaKJJ11}. We will refer to this hypersphere as $\mathcal{C}$, such that, $
\mathcal{C}=\{q: \textit{I} \rightarrow \mathbb{R}^{n} | \; \|q\| = 1\} \subset \mathbb{L}^{2}(\textit{I},\mathbb{R}^{n})\;$. Each element of $\mathcal{C}$ represents a curve in $\mathbb{R}^n$ associated with a human motion. As $\mathcal{C}$ is a hypersphere, the geodesic length between two elements $q_1$ and $q_2$ is defined as: 

\begin{equation}
\label{eq:distance_sphere}
d_{\mathcal{C}}(q_1,q_2)=\cos^{-1}(\langle q_1,q_2 \rangle) \; . 
\end{equation}

\section{Architecture and Loss Functions }\label{sec:predictiveGAN}
Given a set of $m$ action sequences $\{ \{P_1,P_2, \dots P_T \}_i\}_{i=1}^m$ of $T$ consecutive skeleton poses. Let us consider the first $\tau$ poses $(\tau < T)$ as the actions history represented by their corresponding SRVFs $\{q_{\tau}^i\}_{i=1}^m$, and the last $(T-\tau)$ skeleton configurations as the future poses $\{q_{T}^i\}_{i=1}^m$ to be predicted.\\ Motivated by the success of generative adversarial networks, we aim to exploit these generative models to learn an approximation of the function $\Phi : \mathcal{C} \rightarrow \mathcal{C}$ that predicts the $(T-\tau)$ future poses from their associated $\tau$ prior ones. This can be achieved by learning the distribution of SRVFs data corresponding to future poses, on their underlying manifold \emph{i.e.}, hypersphere.  As stated earlier, SRVFs representations are manifold-valued data that cannot be used directly by classical GANs. This is due to the fact that the distribution of data having values on a manifold is quite different from the distribution of those lying on Euclidean space. 
\cite{ZhiwuHuang2017}, exploited the tangent space of the involved manifold and propose a manifold-aware WGAN that generates random data on a manifold. Inspired from this work, we propose a manifold-aware WGAN for motion prediction, to which we refer as PredictiveMA-WGAN, that can predict the future poses from the past ones. This is achieved by using the prior poses as input condition to the MA-WGAN. This condition is also represented by its SRVF; as a result PredictiveMA-WGAN takes manifold-valued data as input to predict its future, which is also a manifold-valued data. 

\subsection{Network Architecture}
PredictiveMA-WGAN consists of two networks trained in an adversarial manner: the predictor $\mathcal{G}$ and the discriminator $\mathcal{D}$. The first network $\mathcal{G}$ adjust its parameters to learn the distribution $\mathbb{P}_{q_T}$ of the future poses $q_T$ conditioned on the input prior ones $q_{\tau} $, while \textit{D} tries to distinguish between the real future poses $q_T$ and the predicted ones $\hat{q}_T$. 
During the training of these networks, we iteratively map the SRVF data back and forth to the tangent space using the exponential and the logarithm maps, defined in a particular point on the hypersphere. 

The predictor network is composed of multiple upsampling and downsampling blocks. It takes as input the prior poses $q_{\tau}$ and output the predicted future poses $\hat{q}_T$. A fully connected layer with $36864$ output channels and five upsampling blocks with $512$, $256$, $128$, $64$ and $1$ output channels, process the input prior pose. These upsampling blocks are composed of the nearest-neighbor upsampling followed by a $3 \times 3$ stride $1$ convolution and a Relu activation. The Discriminator \textit{D} contains three downsampling blocks with $64$, $32$ and $16$ output channels. Each block is a $3 \times 3$ stride $1$ Conv layer followed by batch normalization and Relu activation. These layers are then followed by two fully connected (FC) layers of $1024$ and $1$ outputs. The first FC layer uses Leaky ReLU and batch normalization. 

\subsection{Loss Functions}
In general, the objective of the training consists in minimizing the Wasserstein distance between the distribution of the predicted future poses $\mathbb{P}_{\hat{q}_{T}}$ and that of the real ones $\mathbb{P}_{q_T}$ provided by the dataset. Toward this goal we make use of the following loss functions:

\textbf{Adversarial loss} -- We propose an adversarial loss for predicting manifold-valued data from their history. The predictor takes a manifold-value data $q_{\tau}$ as input rather than a random vector as done in~\cite{ZhiwuHuang2017}, which requires to map these data to a tangent space using the logarithm map before feeding them to the network. Our adversarial loss is the following:
\begin{equation}
\begin{split}
\mathcal{L}_{a}= & {\mathbb{E}_{q_T \sim \mathbb{P}_{q_T}}\left[\mathcal{D}\left(\log _{\mu}(q_T)\right)\right]} 
\\ 
{ } & {-\mathbb{E}_{\mathcal{G}(log_\mu(q_{\tau})) \sim \mathbb{P}_{\hat{q}_T}}\left[\mathcal{D}\left(\log _{\mu}\left(\exp _{\mu}(\mathcal{G}(
\log _{\mu}
(q_{\tau})
))\right)\right)\right]}
\\ 
{ } & {+\lambda \mathbb{E}_{\widetilde{q} \sim \mathbb{P}_{\widetilde{q}}}\left[\left(\left\|\nabla_{\widetilde{q}} \mathcal{D}(\widetilde{q})\right\|-1\right)^{2}\right]},
\end{split}
\end{equation}

where the exponential map, $\exp_\mu(.)$: $T_{\mathcal{\mu}}(C) \mapsto \mathcal{C}$ has a simple expression: $$\exp _{\mu}(s)=cos(\|s\|)\mu + sin(\|s\|)\frac{s}{\|s\|},$$

and the inverse exponential map also called logarithm map $\log _{\mu}(q)$: $\mathcal{C} \mapsto T_{\mathcal{\mu}}(C)$  is given by: 

$$\log _{\mu}(q) = \frac{d_{\mathcal{C}}(q,\mu)}{sin(d_{\mathcal{C}}(q,\mu))} (q - cos(d_{\mathcal{C}}(q,\mu))\mu)$$


where $d_{\mathcal{C}}(.,.)$ is the geodesic distance defined by~\eqref{eq:distance_sphere}.
The last term of $\mathcal{L}_a$ represents the gradient penalty proposed in~\cite{gulrajani2017improved}. $\widetilde{q}$ is a random sample following the distribution $\mathbb{P}_{\widetilde{q}}$, which is sampled uniformly along straight lines between pairs of points sampled from the real distribution $\mathbb{P}_{q_T}$ and the generated distribution $\mathbb{P}_{\hat{q}_T}$. It is given by:
$\label{Eq:eq}
\widetilde{q} = (1 - a) \log_\mu(q_T) + a\log_\mu(\exp_\mu(\mathcal{G}(\log _{\mu}(q_{\tau})))), $ where $\nabla_{\widetilde{q}}  D(\widetilde{q})$  is the gradient with respect to $\widetilde{q}$, and $0\leqslant a \leqslant 1 $.\\
The reference point $\mu$ of the tangent space used in our training is set to the mean of the training data. For a given set of training trajectories $q_1, \dots, q_m$. The mean is given by the Karcher mean~\cite{karcher1977riemannian} in $\mathcal{C}$, 
\begin{equation}
\mu=\underset{{q_i} \in \mathcal{C}}{\operatorname{argmin}} \sum_{i=1}^m d_{\mathcal{C}}^2(\mu, q_i) 
\end{equation}
where $\{q_i \}_{i=1}^m$ is $m$ training data. We present a commonly used algorithm for finding Karcher mean for a given set of curves~\cite{SrivastavaBook2016}. This approach, presented in Algorithm \ref{algo:karcher}.  This computation is based on an  iterative calculation which converges to the optimal solution which is the mean.

\begin{algorithm}
\caption{Karcher mean on $\mathcal{C}$}
\label{algo:karcher}
\
\KwIn {Given SRVFs {$\lbrace q_1,q_2 \cdots q_N \rbrace$,\\ $\epsilon=0.9$, $\tau$: threshold which is a very small number}}
\KwOut {$\mu_j$ : mean of $\lbrace q_i \rbrace _{i=1:N}$}
\BlankLine
\textbf{1-}{ $\mu_0$: initial estimate of Karcher mean, for example one could just take $\mu_0=q_1$, j=0}

\Repeat{$\|\overline{v} \| <\tau$}{

\For{$i\leftarrow 1$ \KwTo $N$}{
\textbf{2-} Compute $v_i=\frac{\theta_i}{sin(\theta_i)}(q_i^* - cos(\theta_i)\mu_j)$, where $cos(\theta_i)=\langle  \mu_j,q_i^*  \rangle$\\
\textbf{3-} Compute the average direction $\overline{v}=\frac{1}{n}\sum_{i=1}^{n} v_i$\\
\textbf{4-} Move $\mu_j$ in the direction of $\overline{v}$ by  $\epsilon$: 
$\mu_{j+1}= cos(\epsilon \| \overline{v}\|)\mu_j+sin(\epsilon \|\overline{v}\|) \frac{\overline{v}}{\| \overline{v} \|}$
}
\textbf{5-} j=j+1
}
\end{algorithm}
\vspace{5mm}

\textbf{Reconstruction loss} -- In order to predict motions close to their ground truth, we add a reconstruction loss $\mathcal{L}_r$. This loss function quantifies the similarities in the tangent space $T_\mu(\mathcal{C})$ between the tangent vector $\log_\mu(q_{T})$ of the ground truth $q_T$ and its associated reconstructed vector $\log_\mu(\exp_\mu(\mathcal{G}(\log _{\mu}(q_{\tau}))))$. It is given by,
\begin{equation}
\label{eq:T_reconstruction_loss}
\mathcal{L}_r = \|{ \log_{\mu}(\exp_\mu({\mathcal{G}(\log _{\mu}(q_{\tau}))})) - \log_\mu(q_{T})} \|_1 \; ,
\end{equation}

\noindent
where $\|.\|_1$ denotes the $L_1$-norm. 

\textbf{Skeleton integrity loss} -- We propose a new loss function  $\mathcal{L}_{s}$ that minimizes the distance between the predicted poses and their ground truth as a remedy to the generation of abnormal skeleton poses.
 Indeed, the aforementioned loss functions rely only on the SRVF representations, which imposes constraints only on the dynamic information. However, to capture the spatial dependencies between joints that avoid implausible poses, we need to impose constraints on the predicted poses directly instead of their motions. By doing so, we predict dynamic changes that fit the initial pose and result in a long-term plausibility. The proposed loss function is based on the Gram matrix of the joint configuration $P$, $G = PP^T$, where $P$ can be seen as $k \times 3$ matrix. Let $G_i, G_j$  be two Gram matrices, obtained from joint poses $P_i, P_j \in \mathbb{R}^{k \times 3}$. The distance between $G_i$ and $G_j$ can be expressed \cite[p.~328]{GoluVanl96} as:
\begin{equation}
\label{eq:distW}
\Delta(G_i,G_j) = \Tr\left(G_i\right) + \Tr\left(G_j\right) - 2 \sum \limits_{{i=1}}^3 \sigma_{i} \; ,
\end{equation} 
\noindent where $\Tr(.)$ denotes the trace operator, and $\{\sigma_i\}_{i=1}^3$ are the singular values of $P_j^T P_i$.
\noindent
The resulting loss function is,

\begin{equation}
\label{eq:loss}
\mathcal{L}_{s} = \dfrac{1}{m}\dfrac{1}{\tau}\sum_{i=1}^{m}\sum_{t=1}^{\tau}\Delta (P_{i,t}, \hat{P}_{i,t})  \; ,
\end{equation}

\noindent where $m$ represents the number of training samples, $\tau$ is the length of the predicted sequence, $P$ is the ground truth pose and $\hat{P}$ is the predicted one.

\textbf{Bone length loss} -- To ensure the realness of the predicted poses, we impose further restrictions on the length of the bones. This is achieved through a loss function that forces the bone length to remain constant over time. Considering $b_{i,j,t}$  and $\hat{b}_{i,j,t}$ the $j$-th bones at time $t$ from the ground truth and the predicted $i$-th skeleton, respectively, we compute the following loss : 
\begin{equation}
\label{eq:loss_bone}
\mathcal{L}_{b} = \dfrac{1}{m}\dfrac{1}{\tau}\dfrac{1}{B}\sum_{i}^{m}\sum_{t=1}^{\tau}\sum_{j}^{B}\norm{b_{i,j,t}-\hat{b}_{i,j,t}}  \; ,
\end{equation}
with B the number of bones in the skeleton representation.

\textbf{Global loss} -- PredictiveMA-WGAN is trained using a weighted sum of the four loss functions $\mathcal{L}_a$, $\mathcal{L}_r$, $\mathcal{L}_s$ and $\mathcal{L}_b$ introduced above, such that,
\begin{equation}
\label{eq:PredictiveGAN_loss} 
\begin{array}{rl}
\mathcal{L}= {\beta_1 \mathcal{L}_{a} + \beta_2 \mathcal{L}_{r} + \beta_3 \mathcal{L}_{s} + \beta_4 \mathcal{L}_{b}}.
\end{array}
\end{equation}
\noindent
The parameters $\beta_i$ are the coefficients associated to different losses, they are set empirically in our experiments. 

The algorithm \ref{alg:WGANalgorithm} summarizes the main steps of our approach. It is divided in two stages, first we outline the steps needed to train our model, then we present the prediction stage, where the trained model is used to predict future poses of a given sequence. 
\begin{algorithm}[!ht]  
\tcp{Training}
  \KwData{$\{q_{\tau}^i\}_{i=1}^m$: SRVFs of training prior poses,
  $\{q_{T}^i\}_{i=1}^m$: real future poses,
  $\theta_0:$ initial parameters of $\mathcal{G}$, 
  $\eta_0:$ initial parameters of $\mathcal{D}$, 
  $\epsilon$: learning rate, $K$: batch size, $\lambda$: balance parameter of gradient penalty, $\zeta$: iterations number.
  }
 \KwResult{$\theta$: generator learned parameters.}
 \nl \For{$i=1 \dots \zeta$}{
 \nl Sample a mini-batch of $K$ random prior poses $\{q^j_{\tau} \}_{j=1}^K \sim \mathbb{P}_{q_{\tau}}$;
 
 \nl Sample a mini-batch of $K$ real future poses; $\{q^j_{T} \}_{j=1}^K \sim \mathbb{P}_{q_T}$;
 
 
 \nl $D_{\eta} \leftarrow{ \Delta_{\eta}(\mathcal{L})}, \mathcal{L}$ is given by Eq.~10;
 
 \nl $\eta \leftarrow \eta + \epsilon . AdamOptimizer(\eta, D_{\eta});$
 
\nl Sample a mini-batch of $K$ random prior poses; $\{q^j_{\tau} \}_{j=1}^K \sim \mathbb{P}_{q_\tau}$;
 
 \nl Compute $\{\mathcal{G}_{\theta}(\log _{\mu}(q_{\tau}^j))\}_{j=1}^K$;
 
\nl $G_{\theta} \leftarrow \Delta_{\theta}( -  D_{\eta}\left(\log _{\mu}\left(\exp _{\mu}(\mathcal{G}_{\theta}(\log _{\mu}(q_{\tau} ))\right)\right) ))$ 
 
\nl $\theta \leftarrow \theta + \epsilon . AdamOptimizer(\theta, G_{\theta});$}

\tcp{Prediction}
 \KwData{$\theta$: generator learned parameters, \\
 \hspace{0.8cm} $ \{ P_i\}_{i=1}^{\tau}$: Prior poses of a testing sequence. }
  \KwResult{$\{ \hat{P}_i\}_{i=\tau+1}^{T}$: Predicted future poses.}
  
 \nl Compute $q_{\tau}$ from $\{ P_i\}_{i=1}^{\tau}$ with Eq.~1;
 
 \nl Compute $\hat{q}_{T}= \exp_{\mu}( \mathcal{G}_{\theta}(\log _{\mu}(q_{\tau} )))$ using the learned parameters $\theta$;
 
 \nl Transform $\hat{q}_{T}$ into pose sequence $\{ \hat{P}_i\}_{i=\tau+1}^{T}$ using Eq.~2, with $\alpha(0) = P_{\tau} $

\caption{PredictiveMAWGAN algorithm}
\label{alg:WGANalgorithm}
\end{algorithm}
\section{Experiments}\label{sec:experiments}

We evaluate the proposed approach with extensive experiments on two popular datasets. In this part we show and discuss our results.

\subsection{Datasets and Pre-processing}

\noindent \textbf{Human 3.6M}~\cite{ionescu_human36m_2014}.  
it is a database that contains 11 subjects performing 15 different actions (Walking, Phoning, Taking photos…). It is one of the largest dataset and the most commonly used for evaluating human motion prediction with 3D skeletons. 
Following the protocol set by previous approaches~\cite{martinez_human_2017,Cui_2020_CVPR} we train our model on $6$ subjects and test it on the specific clips of the $5$th subject.
In the same way as~\cite{Cui_2020_CVPR} out of the $32$ skeletal joints we only use $17$, we remove the joints that correspond to duplicate joints, hands and feet.

 
 For Human3.6M we take the database processed by \cite{jain_structural-rnn_2016} formatted in exponential map and we use their code to convert them to Cartesian coordinates. During our preprocessing step we down sample the sequence  from 50 fps to 25 fps and then perform a normalization by subtracting the mean, dividing by the norm and subtracting the coordinates of the root joint (hips). In the dataset proposed by \cite{jain_structural-rnn_2016} each class of each subject is composed of 2 long sequences. We divide those into smaller sequence for short term prediction (60 frames) and long term prediction (75 frames), following \cite{li_convolutional_2018}. When generating these smaller sequence we avoid overlap, \emph{e.g.} when generating sequence for long term prediction (75 frames) the first sequence contains the frames 1 to 75, the second frames 76 to 150 and so on. This leaves us with 3480 training samples and 812 testing samples for short-term prediction and 2769 training samples and 644 testing samples for long-term prediction.\\


\noindent \textbf{CMU Motion Capture} (CMU MoCap). CMU Mocap dataset 
\footnote{\href{http://mocap.cs.cmu.edu}{http://mocap.cs.cmu.edu}} is a database that contains $5$ categories of motion, each containing several actions. Following~\cite{li_convolutional_2018}, we keep only 8 actions: 'basketball', 'basketball signal', 'directing traffic','jumping', 'running', 'soccer', 'walking' and 'washing window'. We keep the same joint configuration as for Human3.6M and preprocess the data the same way. This leads to 2871 training samples and 704 test samples for short-term prediction and 2825 training samples and 677 test samples for long-term prediction.

\subsection{Implementation Details}

We train our network with a batch size of 64 on 500 epochs and with a learning rate of $10^{-4}$ using the Adam optimizer \cite{KingmaICLR14}. We use $\beta_1=1$, $\beta_2=1$, $\beta_3=10$ and $\beta_4=10$ for the loss coefficients. Our Implementation run on a PC with a Nvidia Quadro RTX 6000 GPU, two 2.3Ghz processors and 64Go of RAM using  Tensorflow 2.2.

\subsection{Evaluation Metrics and Baselines}

We use state-of the art methods for motion prediction that were based on 3D coordinate representation for our comparison. This includes RNN based method (Residual sup).~\cite{martinez_human_2017}, CNN based method (ConvSeq2Seq)~\cite{li_convolutional_2018} and graph models; (FC-GCN) \cite{mao2020learning} and (LDRGCN)~\cite{Cui_2020_CVPR}. 


The zero velocity baseline introduced by ~\cite{martinez_human_2017} is a very simple baseline that use the last observed frame at $t = \tau$ as the value for all the predicted frames, we also compare ourselves to this baseline. The result of LDRGCN are those reported by the authors for the method trained with data in 3D coordinate space. 
Concerning FC-GCN, ConvSeq2Seq and Residual sup., the results are those reported by \cite{mao2020learning} using 3D coordinate data for training. We report the results presented by~\cite{Cui_2020_CVPR} for long-term  prediction (1000ms) results on Human 3.6M, since they are not provided in \cite{mao2020learning}. The long term results for  Residual sup. are not available, we did not include it in our results.


We base our quantitative evaluation on the Mean Per Joint Position Error (MPJPE) \cite{ionescu_human36m_2014} in millimeter following the state-of-the-art \cite{Cui_2020_CVPR}. The metric compare the 3D coordinates of the ground truth with the predicted motions. It is given by,

 \begin{equation}
\label{eq:metric}
MPJPE = \sqrt{\dfrac{1}{\Delta t}\dfrac{1}{k}\sum_{t=\tau+1}^{\tau+\Delta t}\sum_{j=1}^{k}\norm{p_{t,j}-\hat{p}_{t,j}}^2} \; ,
\end{equation}
 where $p_{t,j} = [x_j(t),y_j(t),z_j(t)]$ are the coordinates of joint $j$ at time $t$ from the ground truth sequence, $\hat{p}_{t,j}$ the coordinates from the generated sequence, $k$ the total number of joints in the skeleton, $\tau$ the number of frames in prior sequence and $\Delta t$ the number of predicted frames at which the sequence is evaluated.\\
 
 While MPJPE evaluates the generated samples based on joints positions, it is not enough to assess the evolution of the motion. To complete our assessment we further compare our method with the other approaches based on the evolution along time of the speed of the predicted sequences, we refer to this metric as MPJS (Mean Per Joint Speed). It is computed as follows,
\begin{equation}
\label{eq:metric2}
MPJS(t) = \dfrac{1}{k}\dfrac{1}{M}\sum_{i=1}^{M}\sum_{j=1}^{k}\norm{p_{i,t-1,j}-p_{i,t,j}} \; ,
\end{equation}
with $p_{i,t-1,j}$ and $p_{i,t,j}$ the the coordinates of joint j at time t-1 and t respectively, $k$ the number of joint in the skeleton and M the total number of samples in the test set.

\subsection{Quantitative Comparison}
\subsubsection{Joints position-based evaluation}


To be consistent with recent works, the result are reported for short term prediction and long term prediction. For short term prediction we predict $10$ future frames within $400$ms given $10$ historical frames while we predict $25$ in $1$s based on $25$ prior frames for long term prediction. In Table~\ref{tab:average} we show the comparison of our results with recent methods that use 3D joint coordinates representation. This representation as been proven to provide a more reliable comparison than the angle based representation by ~\cite{mao2020learning}. The results in the table show the clear superiority of our method over methods from the state-of-the-art on both datasets. We highlight that our approach is very competitive with the LDRGN approach for very short term prediction ($80$ms and $160$ms) while outperforming it for longer prediction (320ms, 400ms and 1s). This demonstrate that it is robust when predicting long term motions that stay close to the ground truth.

\begin{table}[!ht]
\centering
\small
\begin{tabular}{@{}l| c c c c c@{} } 
& \multicolumn{5}{c}{Human3.6M average}\\
millisecond (ms) & 80 & 160 & 320 & 400 & 1000 \\ \hline
Zero velocity& 19.6 & 32.5 & 55.1 & 64.4 & 107.9  \cr
Residual sup. & 30.8 & 57.0 & 99.8 & 115.5 & - \cr
convSeq2Seq & 19.6 & 37.8 & 68.1 & 80.3 & 140.5 \cr
FC-GCN & 12.2 & 25.0 & 50.0 & 61.3 & 114.7\cr
LDRGCN&  \textbf{10.7} & \textbf{22.5} & 43.1 & 55.8 & 97.8 \cr
Ours &  12.6 & \textbf{22.5} & \textbf{41.9} & \textbf{50.8} & \textbf{96.4} \cr \hline
& \multicolumn{5}{c}{CMU MoCap average}\\
millisecond (ms) & 80 & 160 & 320 & 400 & 1000 \\ \hline
Zero velocity& 18.4 &31.4  &56.2  & 67.7 &  130.5 \cr
Residual sup. &15.6 & 30.5 & 54.2 & 63.6 & 96.6 \cr
convSeq2Seq &12.5 & 22.2 & 40.7 & 49.7&84.6  \cr
FC-GCN & 11.5 & 20.4 & 37.8 & 46.8 & 96.5\cr 
LDRGCN& \textbf{9.4} & 17.6 & 31.6 & 43.1 & 82.9 \cr
Ours & \textbf{9.4} & \textbf{15.9} & \textbf{29.2} & \textbf{38.3} & \textbf{80.6} \cr \hline
\end{tabular}
\caption{\label{tab:average}Average error over all actions of Human3.6M and CMU MoCap. The short-term in $80$,$160$,$320$,$400ms$, and long-term in $1s$.}
\end{table}

 
In Table~\ref{tab:HUMAN} and~\ref{tab:CMU} we report the results for the literature and for our method on all action classes of Human3.6M and CMU Mocap datasets respectively. The baseline methods adopt a protocol that consist in reporting he average error on eight randomly sampled test sequences. We found that this random sampling can significantly affect the error and makes it hard to present a fair comparison. To avoid this, we decided to report to run the experiment on 8 randomly selected test sequences 100 times, we then report the average error and the standard deviation for these 100 runs for the results of our model. With the standard deviation we can have a better measurement of the general performance of our architecture on different test sequences.
 

 According to Tables~\ref{tab:HUMAN} and~\ref{tab:CMU}, our method perform better than the state-of the art, especially when dealing with long term prediction, these results are consistent with the average error over all actions classes. Interestingly our results also show that the simple zero velocity baseline sometimes  outperforms the state of the art approach on long term prediction (\emph{e.g,} Photo, Sitting and Walking dog for Human3.6H, Soccer and Jumping for CMU MoCap). On the other hand for short term prediction it is always outperformed by the predictions methods. This may be an indication that the MPJPE is not the best suited metric for the problem and a motivation to find a better more representative metric in future works. The results show that the previous approaches performance decrease over time, while ours proves more robust in long term horizons, we are show to perform better than both the zero velocity baseline and the literature. We can notice than some classes present a very large variance  (\emph{e.g,} jumping) while for other the variance is very low (\emph{e.g,} running). This is due to the number of samples which can be be very different from a class to another but also to the high diversity of samples for some classes. Other classes that present less variability  (\emph{e.g,} walking) have a reduced variance.

\begin{table*}[!ht]

\centering
\scriptsize\addtolength{\tabcolsep}{-3pt}
\begin{tabular}{l| p{0.725cm} p{0.725cm} p{0.725cm} p{0.725cm} p{0.725cm}| p{0.725cm} p{0.725cm} p{0.725cm} p{0.725cm} p{0.725cm}| p{0.725cm} p{0.725cm} p{0.725cm} p{0.725cm} p{0.725cm}| p{0.725cm} p{0.725cm} p{0.725cm} p{0.725cm} p{0.725cm} }

& \multicolumn{5}{c|}{Directions} & \multicolumn{5}{c|}{Discussion} & \multicolumn{5}{c|}{Eating} & \multicolumn{5}{c}{Greeting} \cr
millisecond (ms) & 80 & 160 & 320 & 400 & 1000 & 80 & 160 & 320 & 400 & 1000 & 80 & 160 & 320 & 400 & 1000 & 80 & 160 & 320 & 400 & 1000  \cr \hline
Zero velocity & 16.0 & 27.1 & 46.4 & 53.9 & 83.9 & 17.8 & 29.7 & 51.0 & 59.8 & 103.1 & 13.5 & 21.9 & 37.0 & 43.9 & 83.3 & 26.4 & 43.7 & 70.1 & 80.5 & \textbf{124.9}\cr
Residual sup.& 36.5 & 56.4 & 81.5 & 97.3 & - & 31.7 & 61.3 & 96.0 & 103.5 &- & 17.6 & 34.7 & 71.9 & 87.7 & - & 37.9 & 74.1 & 139.0 & 158.8& - \cr
convSeq2Seq& 22.0 & 37.2 & 59.6 & 73.4 & 118.3 & 18.9 & 39.3 & 67.7 & 75.7 & 123.9 & 13.7 & 25.9 & 52.5 & 63.3 & 74.4 & 24.5 & 46.2 & 90.0 & 103.1 & 191.2 \cr
FC-GCN  & 12.6 & 24.4 & 48.2 & 58.4 & 89.1 & 9.8 & 22.1 & 39.6 & 44.1 & 78.5 & 8.8 & 18.9 & 39.4 & 47.2 & 57.1 & 14.5 & 30.5 & 74.2 & 89.0 & 148.4  \cr
LDRGCN & 13.1 & 23.7 & 44.5 & 50.9 & \textbf{78.3} & \textbf{9.4} & \textbf{20.3} & \textbf{35.2} & \textbf{41.2} & \textbf{67.4} & \textbf{7.6} & \textbf{15.9} & 37.2 & 41.7 & \textbf{53.8} & \textbf{9.6} & \textbf{27.9} & 66.3 & 78.8 & 129.7  \cr
Ours & \textbf{11.1} & \textbf{20.9} & \textbf{38.8} & \textbf{47.0} & \textbf{83.5} & 11.9 & \textbf{22.7} & 44.8 & 54.6 & 102.2& \textbf{9.0} & \textbf{15.9} & \textbf{29.1} & \textbf{35.0} & 65.3 & 19.6 & 35.1 & \textbf{64.0} & \textbf{78.2} & \textbf{126.8}\cr
& $\boldsymbol{\std{2.7}}$ & $\boldsymbol{\std{4.9}}$ & $\boldsymbol{\std{8.4}}$ & $\boldsymbol{\std{9.7}}$ & $\boldsymbol{\std{15.3}}$ & $\std{1.9}$ & $\boldsymbol{\std{3.4}}$ & $\std{6.5}$ &$\std{7.7}$ & $\std{16.5}$ & $\boldsymbol{\std{1.5}}$ & $\boldsymbol{\std{2.8}}$ & $\boldsymbol{\std{4.8}}$ & $\boldsymbol{\std{5.3}}$ & $\std{6.8}$ & $\std{3.4}$ & $\std{6.8}$ & $\boldsymbol{\std{13.1}}$ & $\boldsymbol{\std{16.1}}$ & $\boldsymbol{\std{16.7}}$ \cr  \hline \hline

&  \multicolumn{5}{c|}{Phoning} &  \multicolumn{5}{c|}{Photo} & \multicolumn{5}{c|}{Posing} & \multicolumn{5}{c}{Purchase} \cr  
millisecond (ms) & 80 & 160 & 320 & 400 & 1000 & 80 & 160 & 320 & 400 & 1000 & 80 & 160 & 320 & 400 & 1000 & 80 & 160 & 320 & 400 & 1000 \cr \hline
Zero velocity &  15.8 & 26.5 & 43.7 & 51.0 & 92.3 & 16.9 & 28.4 & 49.2 & 58.3 & \textbf{98.8} & 20.4 & 34.7 & 61.5 & 73.3 & 136.1 & 22.1 & 36.5 & 61.8 & 72.2 & 126.3\cr
Residual sup. & 25.6 & 44.4 & 74.0 & 84.2 & - & 23.6 & 47.4 & 94.0 & 112.7 & - & 27.9 & 54.7 & 131.3 & 160.8 & - & 40.8 & 71.8 & 104.2 & 109.8 & - \cr
convSeq2Seq& 17.2 & 29.7 & 53.4 & 61.3 & 127.5 & 14.0 & 27.2 & 53.8 & 66.2 & 151.2 & 16.1 & 35.6 & 86.2 & 105.6 & 163.9 & 29.4 & 54.9 & 82.2 & 93.0 & 139.3  \cr
FC-GCN & 11.5 & 20.2 & 37.9 & 43.2 & 94.3 & \textbf{6.8} & 15.2 & 38.2 & 49.6 & 125.7 & 9.4 & 23.9 & 66.2 & 82.9 & 143.5 & 19.6 & 38.5 & 64.4 & 72.2 & 127.2 \cr
LDRGCN& \textbf{10.4} & \textbf{14.3} & \textbf{33.1} & \textbf{39.7} & 85.8 & 7.1 & \textbf{13.8} & \textbf{29.6} & 44.2 & 116.4 & \textbf{8.7} & \textbf{21.1} & 58.3 & 81.9 & \textbf{133.7} & 16.2  & 36.1 & 62.8 & 76.2 & \textbf{112.6}  \cr 
Ours & \textbf{11.7} & 19.4& \textbf{34.9} & \textbf{42.3} & \textbf{81.8} & \textbf{8.8} & \textbf{16.0} & \textbf{32.4} & \textbf{40.9} & \textbf{98.9} & 13.7 & \textbf{25.9} & \textbf{50.0} & \textbf{61.1} & \textbf{137.7} & \textbf{14.2} & \textbf{26.5} & \textbf{48.3} & \textbf{58.1} & \textbf{120.8}  \cr
& $\boldsymbol{\std{2.2}}$ & $\std{3.6}$ & $\boldsymbol{\std{6.4}}$ & $\boldsymbol{\std{7.6}}$ & $\boldsymbol{\std{9.8}}$ & $\std{2.0}$ & $\boldsymbol{\std{3.5}}$ & $\boldsymbol{\std{6.9}}$ & $\boldsymbol{\std{8.6}}$ & $\boldsymbol{\std{16.1}}$ & $\std{3.3}$ & $\boldsymbol{\std{6.3}}$ & $\boldsymbol{\std{11.0}}$ & $\boldsymbol{\std{12.7}}$ & $\boldsymbol{\std{12.8}}$ & $\boldsymbol{\std{2.5}}$ & $\boldsymbol{\std{4.8}}$ & $\boldsymbol{\std{9.8}}$ & $\boldsymbol{\std{12.5}}$ & $\boldsymbol{\std{19.0}}$ \cr  \hline \hline

& \multicolumn{5}{c|}{Sitting} & \multicolumn{5}{c|}{Sitting Down} & \multicolumn{5}{c|}{Smoking} & \multicolumn{5}{c}{Waiting} \cr   
millisecond (ms) & 80 & 160 & 320 & 400 & 1000 & 80 & 160 & 320 & 400 & 1000 & 80 & 160 & 320 & 400 & 1000 & 80 & 160 & 320 & 400 & 1000 \cr \hline
Zero velocity & 14.6 & 23.9 & 40.9 & 48.4 & \textbf{94.7} & 19.5 & 32.4 & 53.5 & 61.8 & \textbf{112.2} & 14.9 & 24.6 & 41.7 & 49.3 & 84.0 & 17.0 & 28.2 & 48.9 & 57.8 & 99.4\cr
Residual sup.& 34.5 & 69.9& 126.3 & 141.6 & - & 28.6 & 55.3 & 101.6 & 118.9 & - & 19.7 & 36.6 & 61.8 & 73.9 & - & 29.5 & 60.5& 119.9 & 140.6& -\cr
convSeq2Seq& 19.8 & 42.4 & 77.0 & 88.4 & 132.5 & 17.1 & 34.9 & 66.3 & 77.7 & 177.5 & 11.1 & 21.0 & 33.4 & 38.3 & 52.2 & 17.9 & 36.5 & 74.9 & 90.7 & 205.8 \cr
FC-GCN & 10.7 & 24.6 & 50.6 & 62.0 & 119.8 & 11.4 & 27.6 & 56.4 & 67.6 & 163.9 & 7.8 & 14.9 & 25.3 & 28.7 & 44.3 & 9.5 & 22.0 & 57.5 & 73.9 & 157.2 \cr
LDRGCN& \textbf{9.2} & 23.1 & 47.2 & 57.7 & 106.5 & \textbf{9.3} & \textbf{21.4} & \textbf{46.3} & \textbf{59.3} & 144.6 & 8.1 & \textbf{13.4} & \textbf{24.8} & \textbf{24.9} & \textbf{43.1} &  \textbf{9.2} & \textbf{17.6} & 47.2 & 71.6 & 127.3 \cr
Ours & \textbf{10.4}& \textbf{17.9} & \textbf{33.1} & \textbf{40.7} & \textbf{97.7}& 15.8& 28.2& \textbf{52.9} & \textbf{64.5}& \textbf{125.2}& \textbf{7.9} & \textbf{14.3}& \textbf{25.2}& \textbf{30.4}& 63.4& \textbf{11.4} & \textbf{20.3} & \textbf{38.8} & \textbf{47.2} & \textbf{94.0}\cr
& $\boldsymbol{\std{2.8}}$& $\boldsymbol{\std{3.5}}$& $\boldsymbol{\std{5.3}}$& $\boldsymbol{\std{6.4}}$ & $\boldsymbol{\std{14.0}}$ & $\std{3.4}$& $\std{5.1}$ & $\boldsymbol{\std{9.3}}$ & $\boldsymbol{\std{11.5}}$ & $\boldsymbol{\std{23.3}}$ & $\boldsymbol{\std{1.6}}$& $\boldsymbol{\std{2.7}}$ & $\boldsymbol{\std{4.5}}$ & $\boldsymbol{\std{5.2}}$ & $\std{9.7}$ & $\boldsymbol{\std{3.1}}$ & $\boldsymbol{\std{4.3}}$ & $\boldsymbol{\std{7.6}}$ & $\boldsymbol{\std{9.0}}$ & $\boldsymbol{\std{13.7}}$ \cr  \hline \hline

& \multicolumn{5}{c|}{Walking Dog} & \multicolumn{5}{c|}{Walking} & \multicolumn{5}{c|}{Walking Together} & \multicolumn{5}{c}{Average}\cr 
millisecond (ms) & 80 & 160 & 320 & 400 & 1000 & 80 & 160 & 320 & 400 & 1000 & 80 & 160 & 320 & 400 & 1000 &80 &160 & 320 & 400 & 1000 \cr \hline 
Zero velocity & 26.9 & 42.3 & 69.2 & 79.5 & 119.2 & 28.1 & 49.2 & 86.0 & 100.3 & 149.1 & 23.5 & 39.2 & 65.4 & 75.6 & 111.3 & 19.6 & 32.5 & 55.1 & 64.4 & 107.9  \cr
Residual sup.& 60.5 & 101.9 & 160.8 & 188.3 & - & 23.8 & 40.4 & 62.9 & 70.9 & - & 23.5 & 45.0 & 71.3 & 82.8 & - & 30.8 & 57.0 & 99.8 & 115.5 & - \cr
convSeq2Seq& 40.6 & 74.7 & 116.6 & 138.7 & 210.2 & 17.1 & 31.2 & 53.8 & 61.5 & 89.2 & 15.0 & 29.9 & 54.3 & 65.8 & 149.8 & 19.6 & 37.8 & 68.1 & 80.3 & 140.5 \cr
FC-GCN & 32.2 & 58.0 & 102.2 & 122.7 & 185.4 & \textbf{8.9} & 15.7 & 29.2 & 33.4 & 50.9 & 8.9 & 18.4 & 35.3 & 44.3 & 102.4 & 12.2 & 25.0 & 50.0 & 61.3 & 114.7\cr
LDRGCN& 25.3 & 56.6 & 87.9 & 99.4 & 143.2 & \textbf{8.9} & \textbf{14.9} & \textbf{25.4} & \textbf{29.9} & \textbf{45.8} & \textbf{8.2} & \textbf{18.1} & \textbf{31.2} & \textbf{39.4} & 79.2 & \textbf{10.7} & \textbf{22.5} & 43.1 & 55.8 & 97.8 \cr
Ours & \textbf{19.3} & \textbf{34.2} & \textbf{65.6} & \textbf{77.5} & \textbf{117.8} & 12.0 & 21.1& 35.6 & 42.4 & 68.2 & 11.6 & \textbf{19.7} & 34.5 & \textbf{41.8} &\textbf{63.4} & 12.6 & \textbf{22.5} & \textbf{41.9} & \textbf{50.8} & \textbf{96.4} \cr
& $\boldsymbol{\std{5.9}}$ & $\boldsymbol{\std{9.5}}$ & $\boldsymbol{\std{17.8}}$ & $\boldsymbol{\std{19.7}}$ & $\boldsymbol{\std{23.7}}$ & $\std{1.1}$ & $\std{1.7}$ & $\std{2.9}$ & $\std{3.8}$ & $\std{5.3}$ & $\std{1.1}$ & $\boldsymbol{\std{1.6}}$ & $\std{3.0}$ & $\boldsymbol{\std{3.8}}$ & $\boldsymbol{\std{6.4}}$  \cr  \hline

\end{tabular}
\caption{\label{tab:HUMAN}Motion prediction results measured with Equation~\ref{eq:metric} on Human3.6M dataset. Short-term results are reported within $80$, $160$, $320$, $400$ms, and long-term in $1$s. Best results in bold while state-of-the-art best results that fit in our confidence interval are also written bold. }
\end{table*}

\begin{table*}[!ht]

\centering
\scriptsize\addtolength{\tabcolsep}{-3pt}
\begin{tabular}{@{}l| p{0.725cm} p{0.725cm} p{0.725cm} p{0.725cm} p{0.725cm}| p{0.725cm} p{0.725cm} p{0.725cm} p{0.725cm} p{0.725cm}| p{0.725cm} p{0.725cm} p{0.725cm} p{0.725cm} p{0.725cm}| p{0.725cm} p{0.725cm} p{0.725cm} p{0.725cm} p{0.725cm} @{} }

& \multicolumn{5}{c|}{Basketball} & \multicolumn{5}{c|}{Basketball signal} & \multicolumn{5}{c|}{Directing traffic} & \multicolumn{5}{c}{Jumping} \cr
millisecond (ms) & 80 & 160 & 320 & 400 & 1000 & 80 & 160 & 320 & 400 & 1000 & 80 & 160 & 320 & 400 & 1000 & 80 & 160 & 320 & 400 & 1000 \cr \hline
Zero velocity & 20.3 & 34.6 & 62.2& 75.0 & 143.5 & 6.4 & 11.0 & 19.9 & 24.2 & 50.5 & 26.6 & 41.9 & 69.1 & 81.9 &155.3 & 21.4 & 36.3 & 63.2 & 75.2 &138.8 \cr
Residual sup. & 18.4 & 33.8 & 59.5 & 70.5 & 106.7 & 12.7 & 23.8 & 40.3 & 46.7 & 77.5 & 15.2 & 29.6 & 55.1 & 66.1 & 127.1 & 36.0 & 68.7 & 125.0 & 145.5 & 195.5 \cr
convSeq2Seq & 16.7 & 30.5 & 53.8 & 64.3 & \textbf{91.5} & 8.4 & 16.2 & 30.8 & 37.8 & 76.5 & 10.6 & 20.3 & 38.7 & 48.4 & \textbf{115.5} & 22.4 & 44.0 & 87.5 & 106.3 & 162.6 \cr
FC-GCN & 14.0 & 25.4 & 49.6 & 61.4 & 106.1 & 3.5 & 6.1 & 11.7 & 15.2 & 53.9 & 7.4 & \textbf{15.1} & 31.7 & 42.2 & 152.4 & 16.9 & 34.4 & 76.3 & 96.8 & 164.6 \cr
LDRGCN & 13.1 & 22.0 & 37.2 & 55.8 & 97.7 & 3.4 & 6.2 & \textbf{11.2} & \textbf{13.8} & 47.3 & \textbf{6.8} & 16.3 & \textbf{27.9} & \textbf{38.9} & 131.8 & 13.2 & 32.7 & 65.1 & 91.3 & 153.5\cr
Ours & \textbf{9.1} & \textbf{16.6} & \textbf{34.7} & \textbf{44.5} & 108.4 & \textbf{3.3} & \textbf{5.9} & \textbf{11.5} & \textbf{14.7} & \textbf{44.7} & \textbf{19.6} & \textbf{31.3} & \textbf{54.8}& \textbf{66.1}& \textbf{155.5} & \textbf{12.5}& \textbf{22.7} & \textbf{44.4}& \textbf{55.8} & \textbf{120.4}\cr
& $\boldsymbol{\std{0.7}}$ & $\boldsymbol{\std{1.5}}$ & $\boldsymbol{\std{3.5}}$ & $\boldsymbol{\std{4.4}}$ & $\std{5.1}$ & $\boldsymbol{\std{1.1}}$ & $\boldsymbol{\std{2.0}}$ & $\boldsymbol{\std{3.7}}$ & $\boldsymbol{\std{4.7}}$ &$\boldsymbol{\std{15.0}}$& $\boldsymbol{\std{16.7}}$ & $\boldsymbol{\std{23.2}}$ & $\boldsymbol{\std{34.4}}$ & $\boldsymbol{\std{37.4}}$ & $\boldsymbol{\std{52.1}}$& $\boldsymbol{\std{2.0}}$ & $\boldsymbol{\std{3.8}}$ & $\boldsymbol{\std{7.4}}$ & $\boldsymbol{\std{9.6}}$ & $\boldsymbol{\std{21.0}}$    \cr \hline \hline

& \multicolumn{5}{c|}{Running} & \multicolumn{5}{c|}{Soccer} & \multicolumn{5}{c|}{Walking} & \multicolumn{5}{c}{Wash window} \cr
millisecond (ms) & 80 & 160 & 320 & 400 & 1000 & 80 & 160 & 320 & 400 & 1000 & 80 & 160 & 320 & 400 & 1000 & 80 & 160 & 320 & 400 & 1000 \cr \hline
Zero velocity& 30.6& 52.8 & 94.1 & 112.2 &242.6 & 10.3 & 17.5 & 31.8 & 39.0 & 79.4 & 18.3 & 31.2 & 55.1 &66.2 &137.7 &12.3 & 21.1&37.8 &45.7 &90.9\cr
Residual sup. & 15.6 & 19.4 & 31.2 & 36.2 & 43.3 & 20.3 & 39.5 & 71.3 & 84.0 & 129.6 & 8.2 & 13.7 & 21.9 & 24.5 & \textbf{32.2} & 8.4 & 15.8 & 29.3 & 35.4 & 61.1 \cr
convSeq2Seq & 14.3 & \textbf{16.3} & \textbf{18.0} & \textbf{20.2} & \textbf{27.5} & 12.1 & 21.8 & 41.9 & 52.9 & 94.6 & 7.6 & 12.5 & 23.0 & 27.5 & 49.8 & 8.2 & 15.9 & 32.1 & 39.9 & 58.9 \cr
FC-GCN & 25.5 & 36.7 & 39.3 & 39.9 & 58.2 & 11.3 & 21.5 & 44.2 & 55.8 & 117.5 & 7.7 & 11.8 & 19.4 & 23.1 & 40.2 & 5.9 & 11.9 & 30.3 & 40.0 & 79.3 \cr
LDRGCN & 15.2 & 19.7 & 23.3 & 35.8 & 47.4 & 10.3 & 21.1 & 42.7 & 50.9 & 91.4 & \textbf{7.1} & \textbf{10.4} & \textbf{17.8} & \textbf{20.7} & 37.5 & 5.8 & 12.3 & 27.8 & 38.2 & \textbf{56.6} \cr
Ours  & \textbf{12.4}& 19.7 & 32.3 & 39.0 & 68.9 & \textbf{4.9} & \textbf{7.9} & \textbf{14.2} & \textbf{18.0} & \textbf{53.1} & 8.1& 13.6 & 22.1 & 26.1 & \textbf{32.4} &\textbf{5.5} & \textbf{9.8} & \textbf{19.2} & \textbf{24.3} & \textbf{61.3}\cr
& $\boldsymbol{\std{0.0}}$ & $\std{0.0}$ & $\std{0.0}$ & $\std{0.0}$ & $\std{0.0}$ & $\boldsymbol{\std{0.3}}$ & $\boldsymbol{\std{0.6}}$ & $\boldsymbol{\std{1.2}}$ & $\boldsymbol{\std{1.6}}$ & $\boldsymbol{\std{4.5}}$ & $\std{0.0}$ & $\std{0.0}$ & $\std{0.0}$ & $\std{0.0}$ & $\boldsymbol{\std{1.8}}$ & $\std{0.8}$ & $\boldsymbol{\std{1.4}}$ & $\boldsymbol{\std{2.5}}$ & $\boldsymbol{\std{3.1}}$ & $\boldsymbol{\std{8.4}}$ \cr \hline

\end{tabular}
\caption{\label{tab:CMU} Motion prediction results measured with Equation~\ref{eq:metric} on CMU dataset. Short-term results are reported within $80$, $160$, $320$, $400$ms, and long-term in $1$s. Best results in bold while state-of-the-art best results that fit in our confidence interval are also written bold.}
\end{table*}

\subsubsection{Motion-based evaluation}
\begin{figure}
    \centering
    \includegraphics[width=0.9\linewidth]{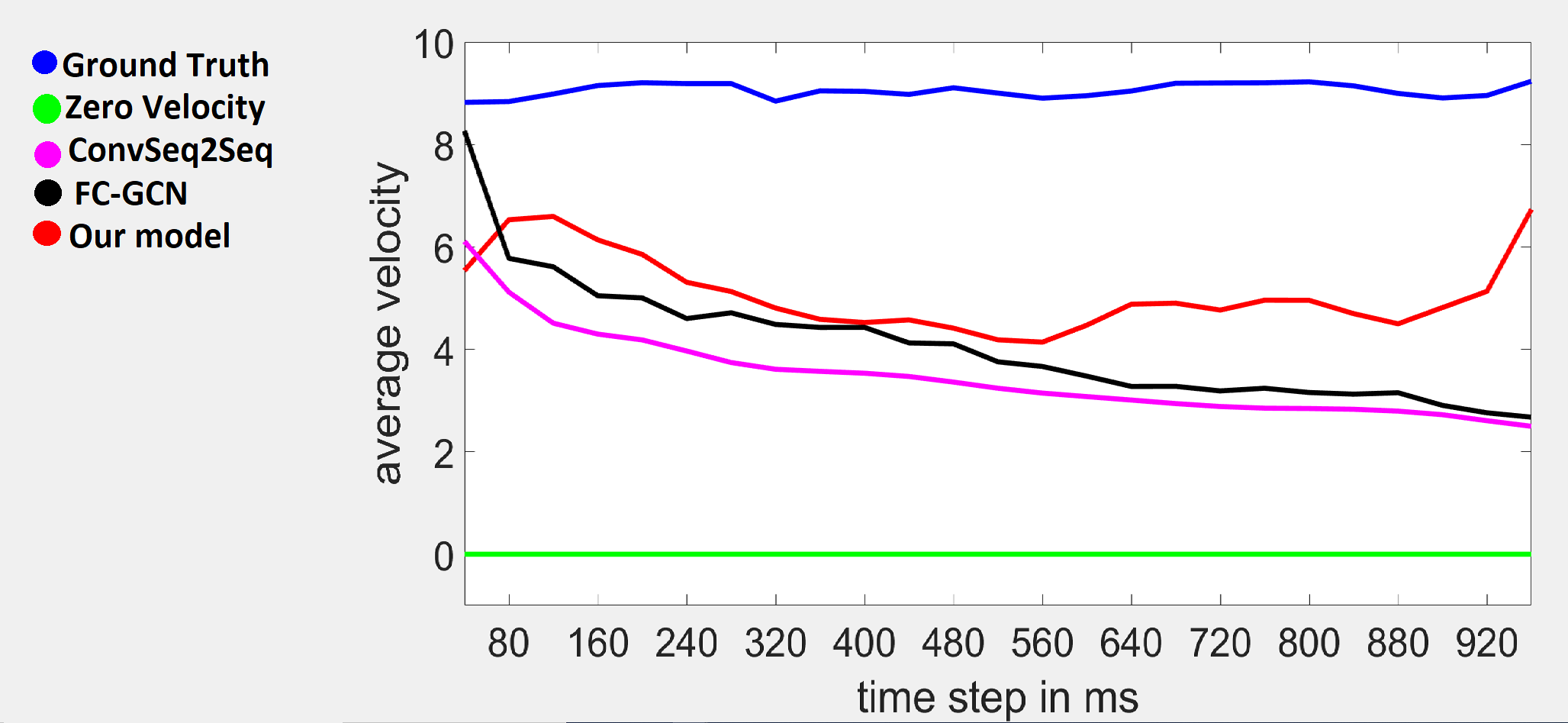}
    \caption{The average speed (MPJS) evolution over 1000 \textit{ms} of all action classes of the Human3.6M dataset. 
    }
    \label{fig:avg_speed}
\end{figure}

\begin{figure}
    \centering
    \includegraphics[width=0.9\linewidth]{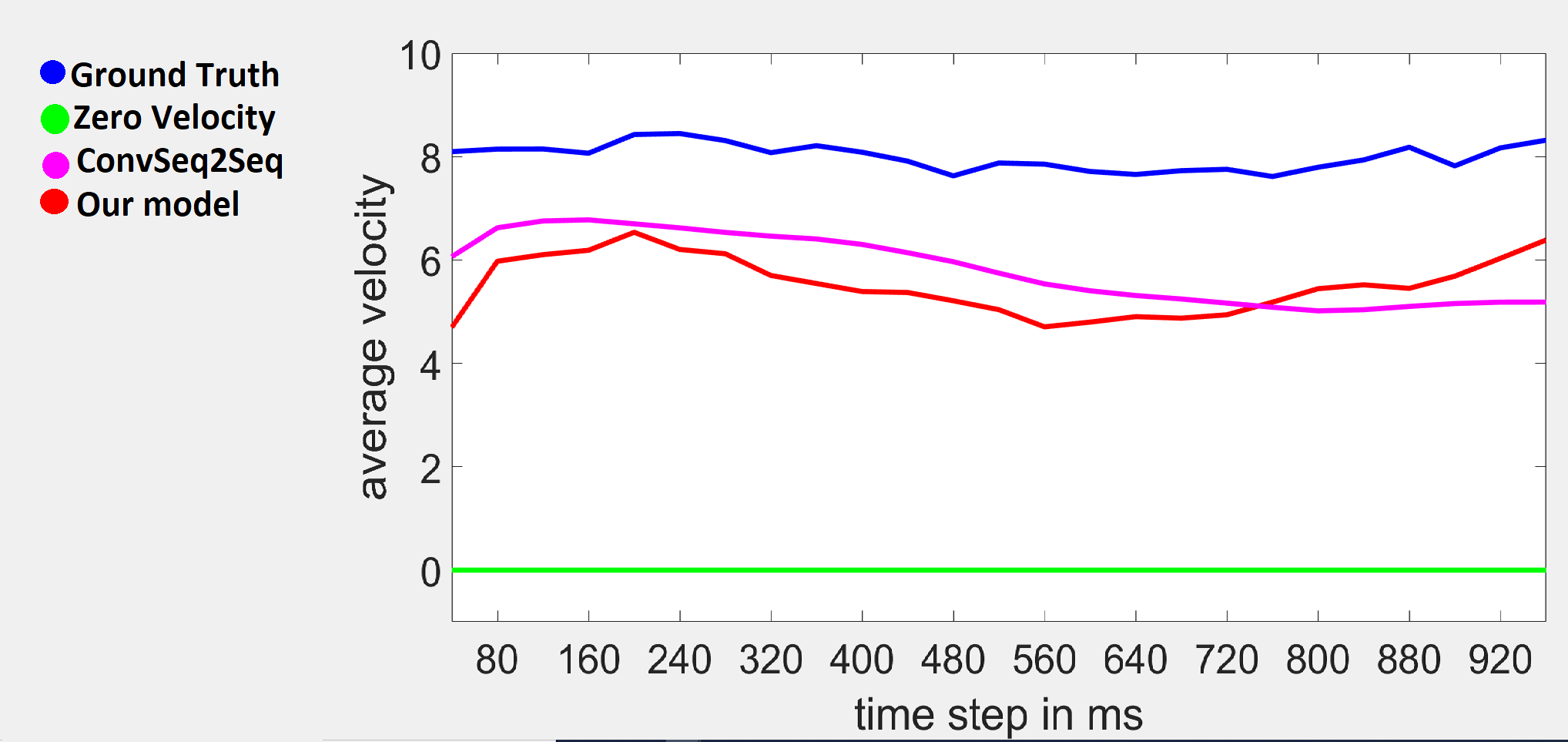}
    \caption{The average speed (MPJS) evolution over 1000 \textit{ms} of all action classes of the CMU MoCap dataset. 
    }
    \label{fig:avg_speed_2}
\end{figure}

To further assess the generated sequences, we evaluate their motion based on the MPJS introduced before. By looking at the evolution of this metric, we can compare our generated motion with the ground truth ones and evaluate the ability of our model to predict motion in long term prediction. 
To this end we show in Figures~\ref{fig:avg_speed} the evolution of the MPJS over time steps on the Human3.6M dataset.\\
Results show that the average ground truth speed vary slightly around 9 \textit{mm/frame}. The zero velocity baseline obviously show the worst results as no motion is being produced and the speed is always null. Both FC-GCN and ConvSeq2Seq speeds continuously decrease over time meaning that less motion is being produced on average for long term horizon. On the other hand our method show variations in the average produced speed up to $1s$ with even an increase during long term prediction. The result from Tables~\ref{tab:HUMAN} and~\ref{tab:average} show that the pose error is good for long time prediction, meaning that the increase of average speed do not correspond to a degradation in the quality of the prediction. This support our claim that our method is a good fit to predict long motion that keep their spatial and temporal coherency. We report in Figure~\ref{fig:avg_speed_2} the evolution of MPJS for the CMU dataset in the same way as for Human3.6M. We do not report the results for FC-GCN as there was no pretrained model available and we were unable to train their model ourselves. The results are similar to those of Human3.6M: average speed of prediction method lower than the ground truth, constant decrease for ConvSeq2Seq, more variation for our method with a significant increase in long term horizons. However we see that this time for short term prediction ConvSeq2Seq produce an average speed closer to the ground truth than ours. This seems to confirm the MPJPE results for ConvSeq2Seq that show that it performs better on CMU than on HUMAN3.6M, we still perform better for long term prediction, showing that our method is robust to different kind of datasets.
Interestingly however all evaluated methods have average speeds well below the values of the ground truth. While this may be explained in part by the presence of sudden, high amplitude and hard to predict motion in action classes like Direction or Greeting, it still indicate that using losses that solely constraint the poses during training lead to generating sequences with slower motion since fast motions are more prone to error. This might be hint that using losses on the speed of the motion will help produce even better predictions.  

We present in Table~\ref{tab:speed} the average MPJS for each class over all time steps on Human3.6M. In consistency with Figure~\ref{fig:avg_speed} this table shows a significant difference between the ground truth and generated motions with all methods. However, this difference changes significantly between classes; some classes like Walking Dog or Greeting present a high difference (6.19 and 4.89 respectively when compared with our method) while other have a lower difference like Eating or Smoking (1.93 and 1.70 respectively when compared with our method). Furthermore, our method is still able to outperform ConvSeq2Seq and FC-GCN on all classes except Walking and Walking together where FC-GCN performs better indicating a capability to better model periodical motion. On the other hand for non-periodic motion our method outperform FC-GCN by a large margin (Greeting, Sitting Down, etc.).


\begin{table}[!ht]
\centering
\small
\begin{adjustbox}{width=\columnwidth,center}
\begin{tabular}{@{}c| c c c c@{} } 
 & Ground truth & ConvSeq2Seq &FC-GCN & Ours\\
 \hline
 Direction & $5.97$ & $2.43$  & $2.39$  &  $3.41$\\
 Discussion & $8.42$ & $3.03$ & $3.28$  & $4.7$\\
 Eating & $6.24$ & $3.35$ &  $3.77$ & $4.31$\\
 Greeting & $11.54$ & $3.41$  &  $3.82$ & $6.65$\\
 Phoning&$7.77$ & $3.29$  &  $3.74$ &$4.66$\\
 Photo&$7.77$ &$2.42$ &  $2.99$ &$3.81$ \\
 Posing&$10.56$ &  $3.34$ &  $3.85$ &$4.84$ \\
 Purchases&$10.28$ &$2.60$    & $3.41$  &$4.97$ \\
 Sitting&$7.37$ &  $1.85$ &  $2.04$ &$3.34$ \\
 Sitting Down &$9.58$ & $2.50$&  $2.37$ &$4.53$ \\
 Smoking&$6.33$ &  $2.90$ &  $4.01$ &$3.95$ \\
 Waiting&$7.98$ & $3.37$  & $3.56$  &$4.63$ \\
 Walking Dog&$13.29$ & $4.59$ &  $5.33$ &$7.1$ \\
 Walking&$12.76$ & $8.11$ & $9.9$  &$8.78$ \\
 Walking together&$9.95$ & $4.95$ & $6.87$  &$6.59$ \\
 \hline
 Average& $9.05$& $3.48$  & $4.09$ &$5.08$ \\
\hline
\end{tabular}
\end{adjustbox}

\caption{\label{tab:speed} Averaged MPJS over 1000 \textit{ms} for all classes of Human3.6M dataset. Closer to the ground truth is better.}
\end{table}

\subsection{Qualitative Comparison}
\begin{figure*}
    \centering
    \includegraphics[width=0.9\linewidth]{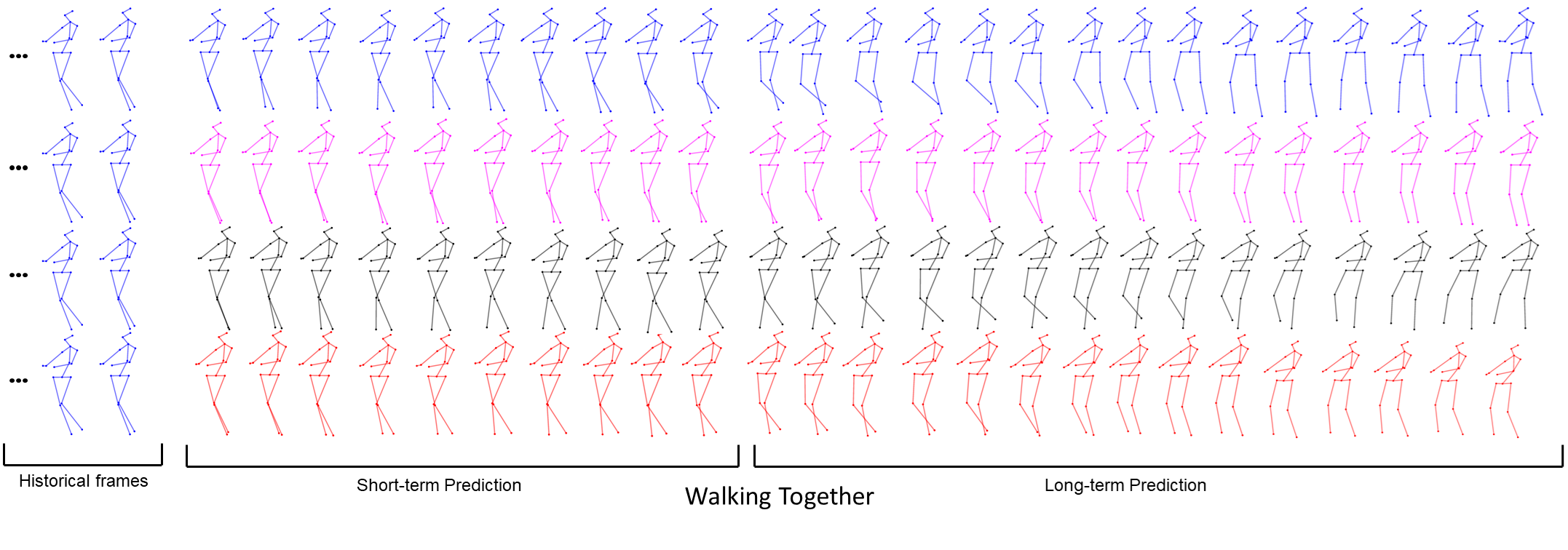}
    \caption{The left frames correspond to the sequence used as a prior. From top to bottom : ground truth, the results of ConvSeq2Seq \protect\cite{li_convolutional_2018}, FC-GCN \protect\cite{mao2020learning} and our method. The illustrated action corresponds to 'Walking Together' from Human3.6M dataset. Short-term frames shown correspond to predicted frames 1, 9 and 10 and long-term frames to frames 11, 12, 22, 23, 24 and 25.}
    \label{fig:introduction}
\end{figure*}

In this part we present some examples that illustrate the smoothness of the generated motion with our method compared to the ground truth and the baselines. \\
In Figure~\ref{fig:introduction} we present the 3D pose sequences of a predicted motion using a model trained for long term prediction with our architecture. We also show the prediction of the same 3D pose sequence by the baseline methods \mbox{ConvSeq2Seq}~\cite{li_convolutional_2018} and FC-GCN~\cite{mao2020learning} using their publicly available code. LDRGCN \cite{Cui_2020_CVPR} is not included as the code for his method is not yet available. We observe that visually our method produce a realistic and smooth motion and that our pose sequence follow more closely the ground truth  than the other methods event for long term prediction. The motion produced by our method do not show any discontinuity, this is the consequence of applying the predicted dynamic of the motion to a starting pose, it prevent the discontinuity than can appear when predicting directly the 3D poses as the other methods do.

\indent

\subsection{Motion Smoothness}

\begin{figure}
    \centering
    \includegraphics[width=0.8\linewidth]{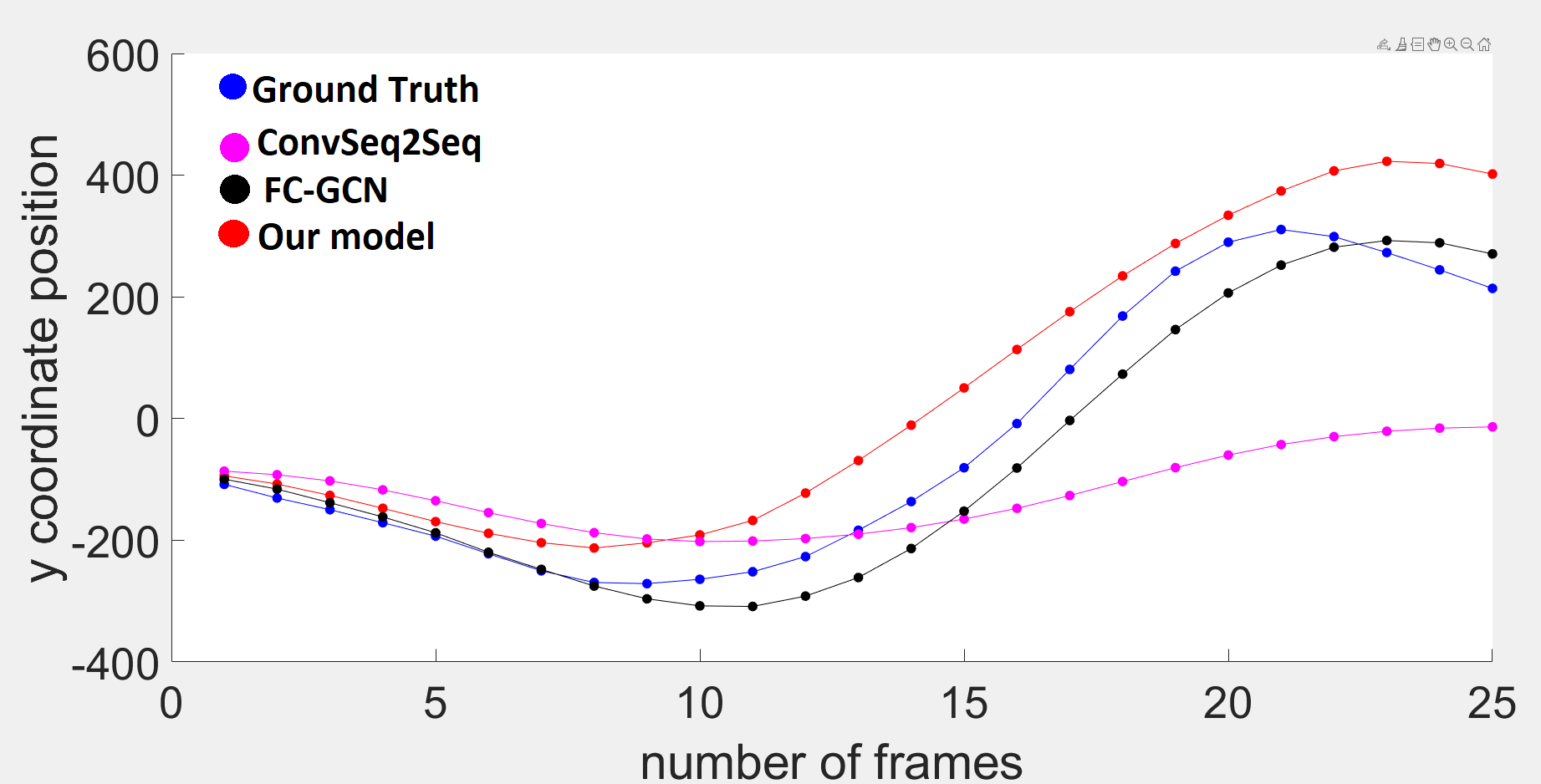}
   \caption{Walking action from Human3.6M.
   X-axis and y-axis corresponds respectively to frame numbers and joint position on the y axis.}
   
    \label{fig:smooth}
\end{figure}

\begin{figure}
    \centering
    \includegraphics[width=0.8\linewidth]{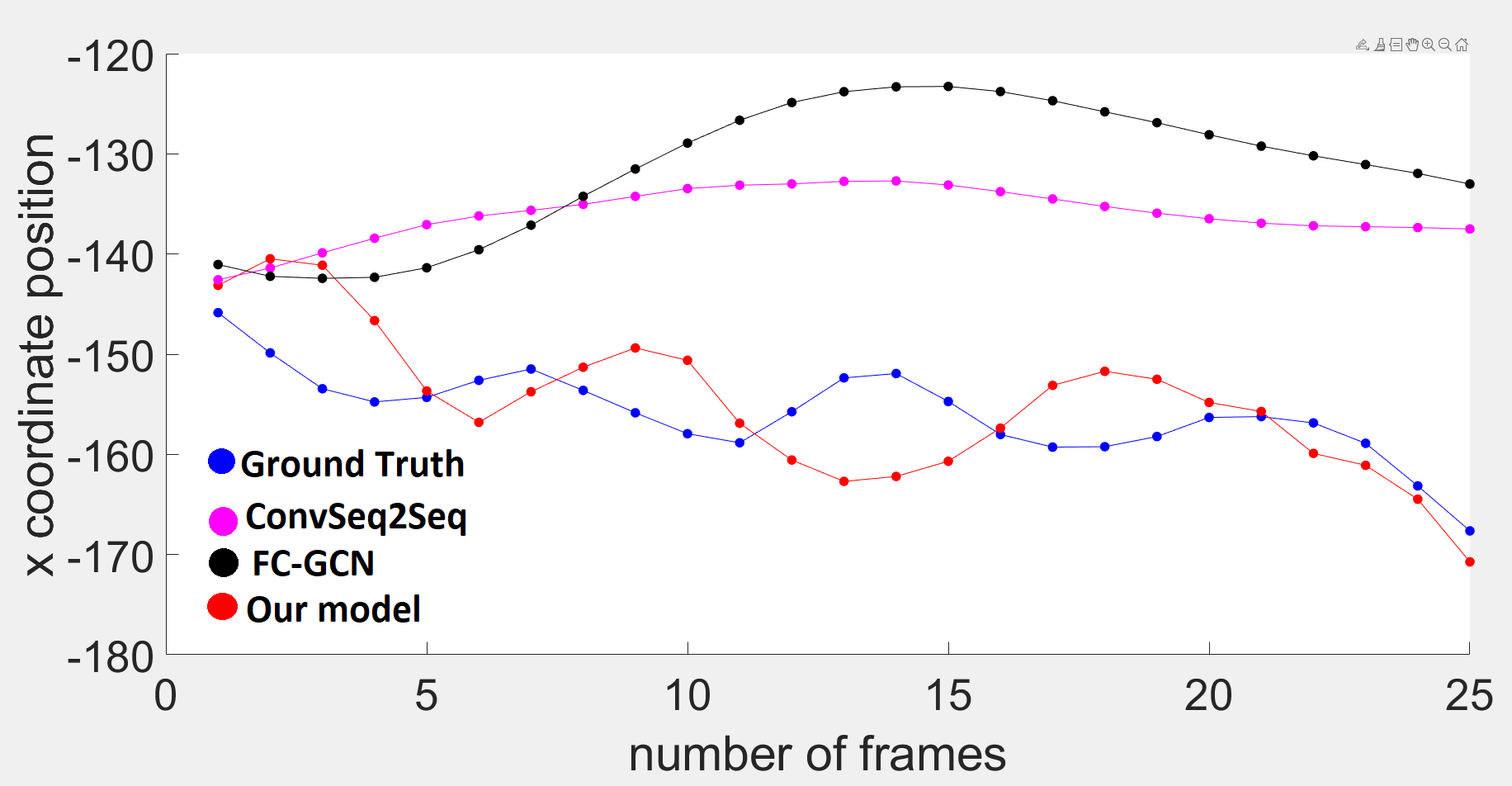}
   \caption{Walking Together action from Human3.6M. 
   X-axis and y-axis corresponds respectively to frame numbers and joint position on the x axis for the right hand joint.}
   
    \label{fig:smooth2}
\end{figure}

In Figure~\ref{fig:smooth} we show the evolution of the y coordinate from the skeleton's left foot over time and in Figure~\ref{fig:smooth2} the evolution of the x axis of the right hand. The 25 frames samples were selected randomly from the walking  and walking together action classes respectively from the Human3.6M dataset. 
We see clearly in the figure that our method is able to generate a smooth motion in both cases and that we are able to follow  the real motion from the ground truth, closely for the walking sample and with a small temporal delay for walking together while for this later, the other methods show a completely different movement.

\subsection{Computation Time}

We show a comparison of the computing time in Table~\ref{tab:time} of our method with \mbox{ConvSeq2Seq} and FC-GCN. This time comparison is done for long term prediction (\emph{i.e}, predicting $25$ frames) with $8$ sequences for each of the 15 action classes from the Human3.6M dataset using the code provided by the author for  \mbox{ConvSeq2Seq} and FC-GCN. The results from Table~\ref{tab:time} show that despite the additional computations required to map the motion back and forth to the tangent space compared to standard GAN architecture, we can predict motion with a speed similar to the other two methods and faster than \mbox{ConvSeq2Seq}.

\begin{table}[!ht]
\centering
\small
\begin{tabular}{@{}c| c c@{} } 

 & total time & time per sample (25 frames)\\
 \hline
 ConvSeq2Seq & $3.04s$ & $\approx25ms$\\
 FC-GCN & $1.67s$ & $\approx14ms$\\
 Ours & $2.42s$ & $\approx20ms$\\
\hline

\hline
\end{tabular}
\caption{\label{tab:time}Prediction time comparison for $8$ predicted samples per action on Human3.6M. }
\end{table}

\subsection{Distribution Visualization}

\begin{figure}[!ht]
\centering
\subfloat[Predicted motions]{
\label{subfig:SNEGroundTruth}
\includegraphics[width=0.4\textwidth]{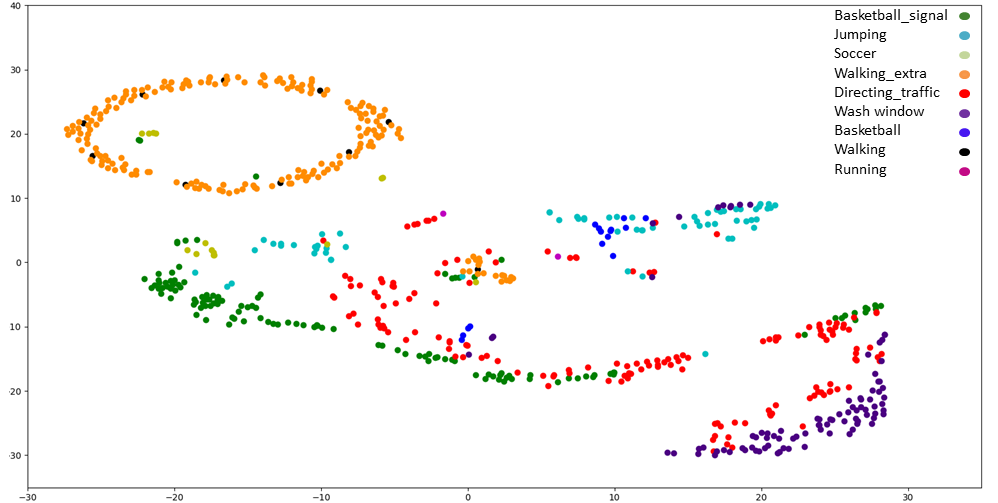}} \\
\subfloat[Ground truth motions]{
\label{subfig:SNEPredictiveMAWGAN}
\includegraphics[width=0.4\textwidth]{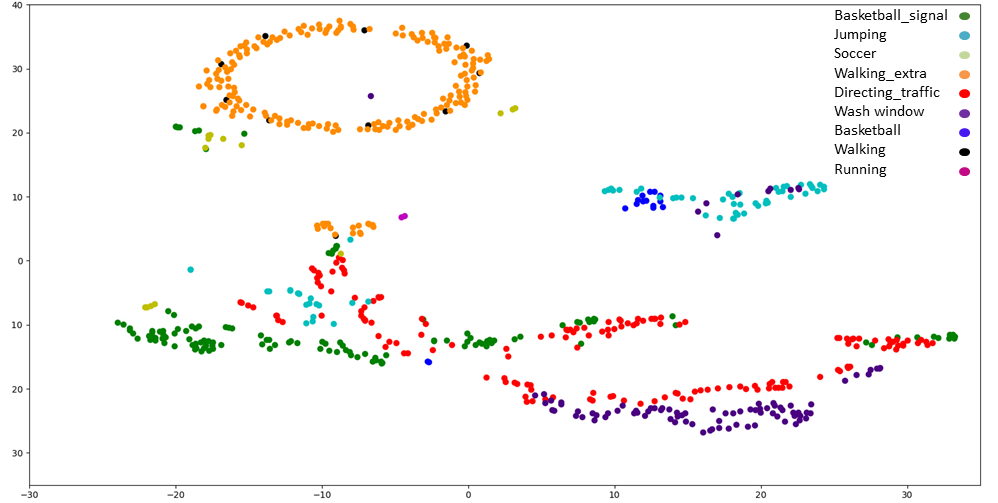} }
\caption{2D visualization of the predicted motions by our method and their associated ground truth using t-SNE algorithm based on Gram distance eq.\ref{eq:distW}. Each color represents an action.}
\label{fig:t-SNE}
\end{figure}

With Figure~\ref{fig:t-SNE} we further assess the quality of the predicted samples using the t-Distributed Stochastic Neighbor Embedding (t-SNE) algorithm~\cite{vandermaaten08a}. We present a 2D visualization of 677 samples of long term prediction from the CMU MoCap dataset. The resulting representation clearly indicates that the motion from the ground truth and the predicted motion are from very close distributions. Moreover we can see that the different generated 3D sequences from the same action are relatively distant from each other, meaning for the same action class our model can predict several motions while respecting the prior motion used for the prediction. 

\subsection{Recursive Generation}
\begin{figure}
    \centering
    \includegraphics[width=0.9\linewidth]{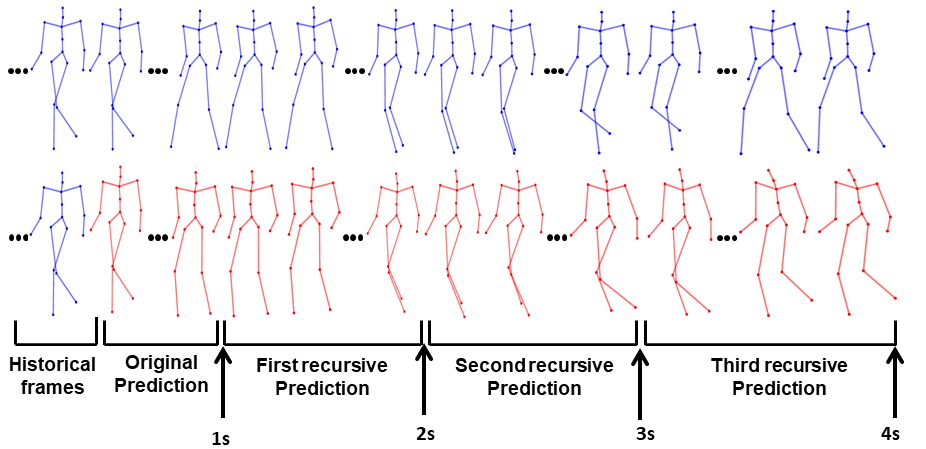}
    \caption{Exemple of recursive prediction on sample of action class walking from the CMU MoCap Dataset for a total of 4 seconds of prediction. On top the ground truth, on the bottom our prediction. }
    \label{fig:recursive}
\end{figure}

 One of the main limitation of our method is its inability to generate sequence of lengths it has not been trained on. We can however still generate longer sequences trough recursive generation by predicting subsequent motion based on previous prediction. 
 This recursive generation can be done without specific training simply by modifying the input during testing. However, by feeding our prediction to the network to get further prediction we cause the network to accumulate error over each recursive iteration. In fact we can not reliably extend the duration of the prediction more than 2 or 3 times. For all types of motion the first and second prediction using predicted data as input are good, the third one is usually still good for periodic motion (\emph{e.g} walking) but not for non-periodic motion (\emph{e.g} greeting). From the fourth prediction onward even the periodic motion will start to deteriorate significantly. Non-periodic motion will usually freeze into a static pose which is to be expected as our prediction can only predict the end of the motion, not infer what other motion might follow. For periodic motion the deterioration comes from the accumulating error which will cause the skeleton to deform. Still we are able to generate motion 3 to 4 time longer than what the network was trained on, which allows us to tackle one of the limitation of the method in some ways. We present in Figure~\ref{fig:recursive} an example of recursive prediction on the action class walking from the CMU MoCap dataset. The figure shows the original prediction and the three subsequent predictions. This shows that our model can predict motion for longer sequences than it has been trained on as we observe significant differences only during the third recursive prediction, the motion however still follow walking action.
 
\subsection{Ablation Study}

\begin{figure}
    \centering
    \includegraphics[width=0.9\linewidth]{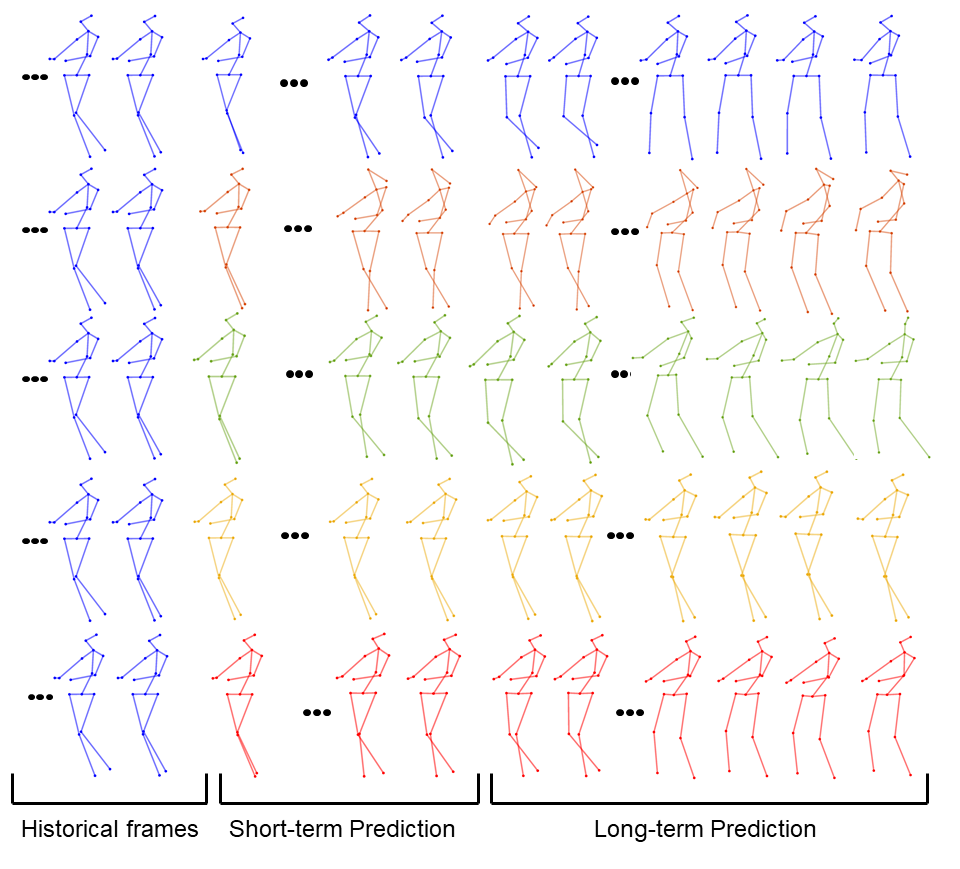}
    \caption{Impact of the bone length loss and the skeleton integrity loss on prediction quality on a sample from action class Walking together from Human3.6M. From top to bottom: the ground truth, neither $\mathcal{L}_s$ nor $\mathcal{L}_b$, only $\mathcal{L}_s$, only $\mathcal{L}_b$ and both $\mathcal{L}_s$ and $\mathcal{L}_b$ }
    \label{fig:ablation}
\end{figure}

To show the efficiency of the different losses used by our network especially the effect of the combination of the skeleton integrity loss $\mathcal{L}_s$ and the bone length loss $\mathcal{L}_b$, we perform our ablation study using model that were trained using only the mentioned losses. The ablation is performed on the Human3.6M dataset due to the huge quantity of data from the dataset. The ablation results are reported in Table~\ref{tab:ablation} for short term and long term prediction using the the average error of all actions classes at different time steps. The results show a clear improvement when adding one of either the skeleton integrity loss or the bone length loss compared to using only $\mathcal{L}_a$ and $\mathcal{L}_r$. Furthermore using both $\mathcal{L}_s$ and $\mathcal{L}_b$ improve significantly the results for long term prediction while keeping a similar accuracy for short term prediction with regard to using only $\mathcal{L}_s$ or only $\mathcal{L}_b$. This evidences the importance of using both losses when doing long term prediction, it allows the model to capture the spatial dependencies between joints and to be able to predict plausible poses even for longer term horizons. \\
We show in Figure~\ref{fig:ablation} the effect of the losses on the visual quality of the prediction.
We notice that excluding $\mathcal{L}_s$ and $\mathcal{L}_b$ leads to important deformations in the upper body but the produced legs motion is rather coherent. Adding $\mathcal{L}_s$ helps produce a motion closer to the ground truth, we however still see noticeable bones deformations (better seen as animations in the supplementary material ) even if we are able to keep a coherent skeleton. Using only $\mathcal{L}_b$ leads to a skeleton without any deformation even during long time prediction but also to very little motion being produced. Using both losses allows us to keep the best skeleton coherency while producing a motion that is close to that of the ground truth.

\begin{table}[!ht]
\centering
\small
\begin{tabular}{@{}c| c c c c c@{} } 
\hline
loss functions & 80 & 160 & 320 & 400 & 1000 \\
\hline
$\mathcal{L}_a + \mathcal{L}_r$ &	20.2 &	34.9 & 	62.4 &	74.9 & 133.3\\
$\mathcal{L}_a + \mathcal{L}_r + \mathcal{L}_s$ & 13.6 &23.4& 42.6 & 51.6 & 103.8 \\
$\mathcal{L}_a + \mathcal{L}_r + \mathcal{L}_b$ & 12.6 &22.4 & 	\textbf{41.3} & 	\textbf{49.9} & 105.6\\
$\mathcal{L}_a + \mathcal{L}_r + \mathcal{L}_s + \mathcal{L}_b$ &\textbf{12.3} &	\textbf{22.2} &	\textbf{41.3} &	50.1 & \textbf{96.2} \\
 \hline
 \end{tabular}
\caption{\label{tab:ablation} Impact of the bone length loss and the skeleton integrity loss on the prediction performance for short-term and long-term. }

\end{table}

\section{Conclusion and Limitations}\label{sec:Conclusion}

In this paper we presented a new and robust method to deal with human motion prediction. In our method we represent the temporal evolution of 3D human poses as trajectories, these trajectories can  be mapped to points on a hypersphere. To be able to learn learn this manifold-valued representation we use a manifold-aware Wasserstein GAN that can capture both the spatial and temporal dependencies involved in human motion. Through extensive experiments we prove the robustness of our method for long term motion prediction when compared to recent literature. With our qualitative results we confirm that we are able to predict plausible poses and smooth motions in long term horizons. \\
The two main limitations of the proposed method are the following: the fixed length of sequence and the inability to deal with sudden changes in motion.
The fixed length in motion is a consequence of the GAN architecture where the input and output sizes a fixed. We demonstrate in our experiment that we can deal with this problem by using recursive generation that show the ability of the model to generate up to 4\textit{s} motion for some classes when trained on 1\textit{s} sequences. 
The inability to deal with a sudden change of motion is inherent to the way motion prediction is usually approached. Indeed, we only consider the historical motion as a condition to predict the motion but it is not always enough to get an accurate prediction. Things like the environment, the goal of the motion and the motion of other persons can influence the future. Taking some of these modalities into account would surely allows for longer and more accurate predictions.



%



\ifCLASSOPTIONcompsoc
  \section*{Acknowledgments}
\else
  \section*{Acknowledgment}
\fi

This project has received financial support from the CNRS through the 80—Prime program and from the French State, managed by the National Agency for Research (ANR) under the Investments for the future program with reference ANR- 16-IDEX-0004 ULNE.

\ifCLASSOPTIONcaptionsoff
  \newpage
\fi

\vskip -2\baselineskip plus -1fil

\begin{IEEEbiography}[{\includegraphics[width=1in,height=1.25in,clip,keepaspectratio]{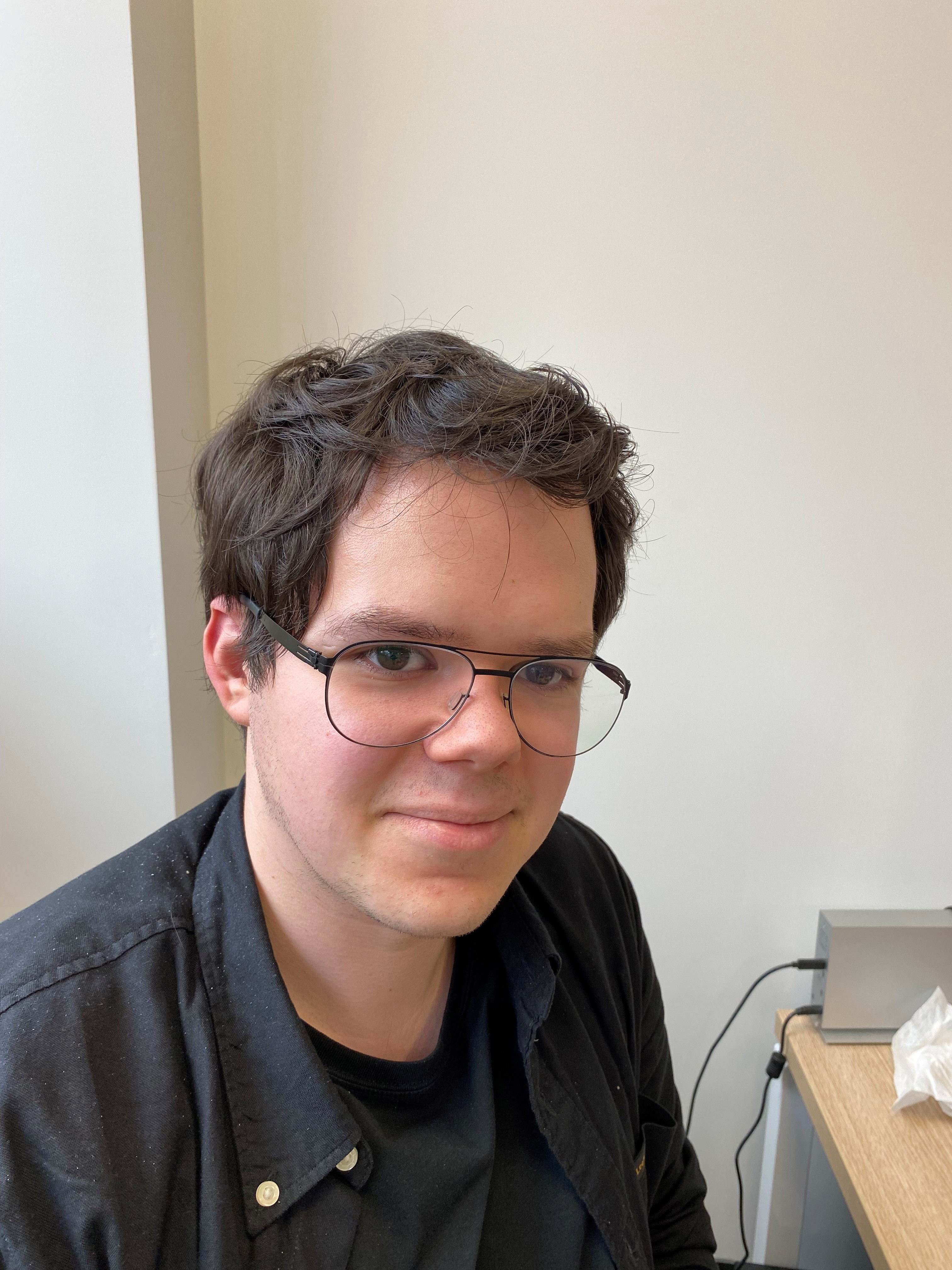}}]
{Baptiste Chopin} received the Engineering degree in computer science from IMT Nord Europe (France).  He is currently pursuing the Ph.D. degree with the university of Lille (France). His research concern computer vision and the generation of human motion with application to cognitive sciences.
\end{IEEEbiography}

\vskip -2\baselineskip plus -1fil

\begin{IEEEbiography}[{\includegraphics[width=1in,height=1.25in,clip,keepaspectratio]{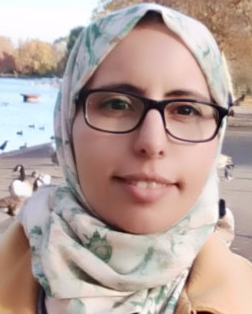}}]
{Naima Otberdout} received the master’s degree in computer sciences and telecommunication
from Mohammed V University, Rabat, Morocco in 2016. She received the Ph.D. degree in computer science from the same university in 2021. She is currently a post-doctoral researcher in the University of Lille, Lille, France. 
\\ Her current research interests include computer vision and pattern recognition with applications to human behavior understanding.
\end{IEEEbiography}

\vskip -2\baselineskip plus -1fil

\begin{IEEEbiography}[{\includegraphics[width=1in,height=1.25in,clip,keepaspectratio]{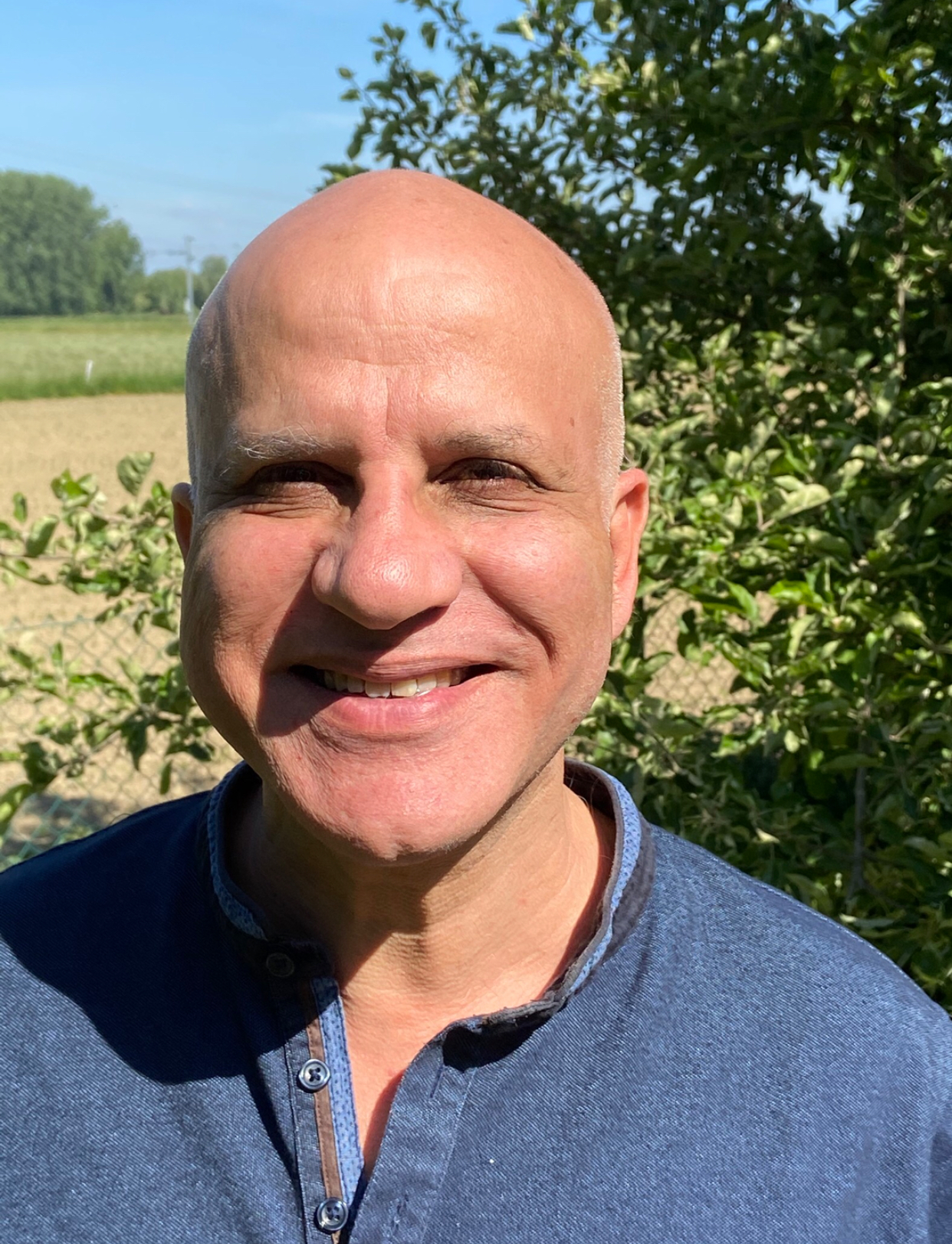}}] {Mohamed Daoudi} is Full Professor of Computer Science at IMT Nord Europe and the lead of Image group at CRIStAL Laboratory (UMR CNRS 9189). He received his Ph.D. degree in Computer Engineering from the University of Lille (France) in 1993. His research interests include pattern recognition, shape analysis and computer vision. He has published over 150 papers in some of the most distinguished scientific journals and international conferences. He is Associate Editor of Image and Vision Computing Journal, IEEE Trans. on Multimedia, Computer Vision and Image Understanding, IEEE Trans. on Affective Computing and Journal of Imaging. He has served as General Chair of IEEE International Conference on Automatic Face and Gesture Recognition, 2019. He is Fellow of IAPR and IEEE Senior member.
\end{IEEEbiography}

\vskip -2\baselineskip plus -1fil

\begin{IEEEbiography}[{\includegraphics[width=1in,height=1.25in,clip,keepaspectratio]{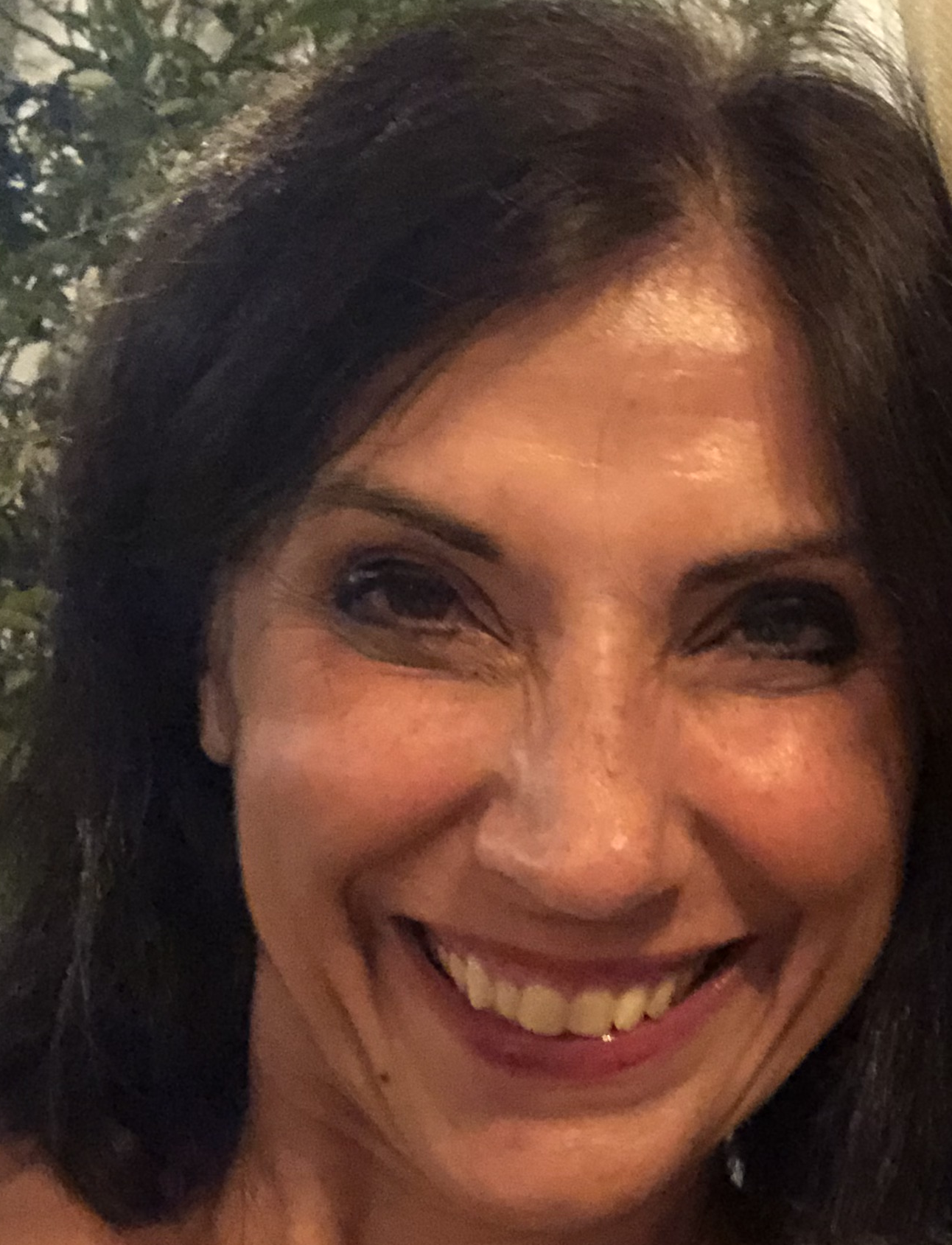}}]
{Angela BARTOLO} is Full Professor of Neuropsychology of Motor Cognition at the University of Lille and Dean of the Faculty of Psychology, Educational and Formation Sciences of the same university. She is co-responsible of the European Master in Psychology of Neurocognitive Processes and Affective Sciences. 
She received her Ph.D. degree in Science from the University of Aberdeen (UK) in 2002 and  was subsequently recruited as post doc researcher at the University of Toronto (Canada), at the University of Modena and Reggio Emilia (Italy) and at the University of Lille (France).
She is member of the Laboratory SCALab (UMR CNRS 9193) and scientific coordinator of the ESTRA International Association Laboratory. Her research interests focus on the processing of manual actions, action semantic and on the relation between action and social cognition in normal population and in brain damage patients, these issues are investigated by means of behavioural and neuroimgaing techniques. She has published over 70 scientific outputs in some of the most distinguished scientific journals. She is a former Junior Member of the Institut Universitaire de France (IUF).
\end{IEEEbiography}






\begin{thebibliography}{10}
\providecommand{\url}[1]{#1}
\csname url@samestyle\endcsname
\providecommand{\newblock}{\relax}
\providecommand{\bibinfo}[2]{#2}
\providecommand{\BIBentrySTDinterwordspacing}{\spaceskip=0pt\relax}
\providecommand{\BIBentryALTinterwordstretchfactor}{4}
\providecommand{\BIBentryALTinterwordspacing}{\spaceskip=\fontdimen2\font plus
\BIBentryALTinterwordstretchfactor\fontdimen3\font minus
  \fontdimen4\font\relax}
\providecommand{\BIBforeignlanguage}[2]{{%
\expandafter\ifx\csname l@#1\endcsname\relax
\typeout{** WARNING: IEEEtran.bst: No hyphenation pattern has been}%
\typeout{** loaded for the language `#1'. Using the pattern for}%
\typeout{** the default language instead.}%
\else
\language=\csname l@#1\endcsname
\fi
#2}}
\providecommand{\BIBdecl}{\relax}
\BIBdecl

\bibitem{KoppulaIEEEROS2013}
H.~S. {Koppula} and A.~{Saxena}, ``Anticipating human activities for reactive
  robotic response,'' in \emph{IROS}, 2013, pp. 2071--2071.

\bibitem{DBLP:journals/tiv/PadenCYYF16}
B.~Paden, M.~C{\'{a}}p, S.~Z. Yong, D.~S. Yershov, and E.~Frazzoli, ``A survey
  of motion planning and control techniques for self-driving urban vehicles,''
  \emph{T-IV}, vol.~1, no.~1, pp. 33--55, 2016.

\bibitem{kovar2008motion}
L.~Kovar, M.~Gleicher, and F.~H. Pighin, ``Motion graphs,'' in \emph{{ACM}
  {SIGGRAPH} Classes}, 2008, pp. 51:1--51:10.

\bibitem{DBLP:conf/iccv/FragkiadakiLFM15}
K.~Fragkiadaki, S.~Levine, P.~Felsen, and J.~Malik, ``Recurrent network models
  for human dynamics,'' in \emph{ICCV}, 2015, pp. 4346--4354.

\bibitem{jain_structural-rnn_2016}
A.~Jain, A.~R. Zamir, S.~Savarese, and A.~Saxena, ``Structural-{RNN}: Deep
  learning on spatio-temporal graphs,'' in \emph{CVPR}, 2016, pp. 5308--5317.

\bibitem{ghosh2017learning}
P.~Ghosh, J.~Song, E.~Aksan, and O.~Hilliges, ``Learning human motion models
  for long-term predictions,'' in \emph{3DV}, 2017, pp. 458--466.

\bibitem{martinez_human_2017}
J.~Martinez, M.~J. Black, and J.~Romero, ``On human motion prediction using
  recurrent neural networks,'' in \emph{CVPR}, 2017, pp. 4674--4683.

\bibitem{li_convolutional_2018}
C.~Li, Z.~Zhang, W.~S. Lee, and G.~H. Lee, ``Convolutional {Sequence} to
  {Sequence} {Model} for {Human} {Dynamics},'' in \emph{CVPR}, 2018, pp.
  5226--5234.

\bibitem{mao2020learning}
M.~Wei, L.~Miaomiao, S.~Mathieu, and L.~Hongdong, ``Learning trajectory
  dependencies for human motion prediction,'' in \emph{ICCV}, 2019, pp.
  9488--9496.

\bibitem{butepage2017deep}
J.~Butepage, M.~J. Black, D.~Kragic, and H.~Kjellstrom, ``Deep representation
  learning for human motion prediction and classification,'' in \emph{CVPR},
  2017, pp. 6158--6166.

\bibitem{MaoICCV19}
W.~Mao, M.~Liu, M.~Salzmann, and H.~Li, ``Learning trajectory dependencies for
  human motion prediction,'' in \emph{ICCV}, 2019, pp. 9488--9496.

\bibitem{BerrettiACMTOM}
\BIBentryALTinterwordspacing
S.~Berretti, M.~Daoudi, P.~K. Turaga, and A.~Basu, ``Representation, analysis,
  and recognition of 3d humans: {A} survey,'' \emph{{ACM} Trans. Multim.
  Comput. Commun. Appl.}, vol.~14, no.~1s, pp. 16:1--16:36, 2018. [Online].
  Available: \url{https://doi.org/10.1145/3182179}
\BIBentrySTDinterwordspacing

\bibitem{chopin2021human}
B.~Chopin, N.~Otberdout, M.~Daoudi, and A.~Bartolo, ``Human motion prediction
  using manifold-aware wasserstein gan,'' in \emph{2021 16th IEEE International
  Conference on Automatic Face and Gesture Recognition (FG 2021)}.\hskip 1em
  plus 0.5em minus 0.4em\relax IEEE, 2021, pp. 1--8.

\bibitem{ferrari_adversarial_2018}
L.-Y. Gui, Y.-X. Wang, X.~Liang, and J.~M.~F. Moura, ``Adversarial
  {Geometry}-{Aware} {Human} {Motion} {Prediction},'' in \emph{ECCV}, 2018, pp.
  823--842.

\bibitem{barsoum2018hp}
E.~Barsoum, J.~Kender, and Z.~Liu, ``Hp-gan: Probabilistic 3{D} human motion
  prediction via gan,'' in \emph{CVPR Workshops}, 2018, pp. 1418--1427.

\bibitem{ZhiwuHuang2017}
Z.~Huang, J.~Wu, and L.~V. Gool, ``Manifold-valued image generation with
  {W}asserstein generative adversarial nets,'' in \emph{{AAAI}}, 2019, pp.
  3886--3893.

\bibitem{Turagacvpr2009}
P.~K. Turaga and R.~Chellappa, ``Locally time-invariant models of human
  activities using trajectories on the {G}rassmannian,'' in \emph{CVPR}, 2009,
  pp. 2435--2441.

\bibitem{Boulbaba2016PAMI}
B.~{Ben Amor}, J.~Su, and A.~Srivastava, ``Action recognition using
  rate-invariant analysis of skeletal shape trajectories,'' \emph{PAMI},
  vol.~38, no.~1, pp. 1--13, 2016.

\bibitem{KacemPAMI2020}
\BIBentryALTinterwordspacing
A.~Kacem, M.~Daoudi, B.~{Ben Amor}, S.~Berretti, and J.~C. {{\'{A}}lvarez
  Paiva}, ``A novel geometric framework on {Gram} matrix trajectories for human
  behavior understanding,'' \emph{PAMI}, vol.~42, no.~1, pp. 1--14, 2020.
  [Online]. Available: \url{https://doi.org/10.1109/TPAMI.2018.2872564}
\BIBentrySTDinterwordspacing

\bibitem{devanne20143}
M.~Devanne, H.~Wannous, S.~Berretti, P.~Pala, M.~Daoudi, and A.~Del~Bimbo,
  ``3-{D} human action recognition by shape analysis of motion trajectories on
  {R}iemannian manifold,'' \emph{IEEE TC}, vol.~45, no.~7, pp. 1340--1352,
  2014.

\bibitem{SrivastavaKJJ11}
\BIBentryALTinterwordspacing
A.~Srivastava, E.~Klassen, S.~H. Joshi, and I.~H. Jermyn, ``Shape analysis of
  elastic curves in euclidean spaces,'' \emph{PAMI}, vol.~33, no.~7, pp.
  1415--1428, 2011. [Online]. Available:
  \url{https://doi.org/10.1109/TPAMI.2010.184}
\BIBentrySTDinterwordspacing

\bibitem{drira20133d}
H.~Drira, B.~{Ben Amor}, A.~Srivastava, M.~Daoudi, and R.~Slama, ``3{D} face
  recognition under expressions, occlusions, and pose variations,''
  \emph{PAMI}, vol.~35, no.~9, pp. 2270--2283, 2013.

\bibitem{OtberdoutPAMI2020}
N.~Otberdout, M.~Daoudi, A.~Kacem, L.~Ballihi, and S.~Berretti, ``Dynamic
  facial expression generation on hilbert hypersphere with conditional
  {W}asserstein generative adversarial nets,'' \emph{PAMI}, pp. 1--1, 2020.

\bibitem{gulrajani2017improved}
I.~Gulrajani, F.~Ahmed, M.~Arjovsky, V.~Dumoulin, and A.~C. Courville,
  ``Improved training of {W}asserstein {GAN}s,'' in \emph{NIPS}, 2017, pp.
  5767--5777.

\bibitem{karcher1977riemannian}
H.~Karcher, ``Riemannian center of mass and mollifier smoothing,''
  \emph{Communications on pure and applied mathematics}, vol.~30, no.~5, pp.
  509--541, 1977.

\bibitem{SrivastavaBook2016}
A.~Srivastava and E.~P. Klassen, \emph{Functional and Shape Data
  Analysis}.\hskip 1em plus 0.5em minus 0.4em\relax Springer, New York, NY,
  2016.

\bibitem{GoluVanl96}
G.~H. Golub and C.~F. Van~Loan, \emph{Matrix Computations}, 4th~ed.\hskip 1em
  plus 0.5em minus 0.4em\relax The Johns Hopkins University Press, 1996.

\bibitem{ionescu_human36m_2014}
\BIBentryALTinterwordspacing
C.~Ionescu, D.~Papava, V.~Olaru, and C.~Sminchisescu,
  ``\BIBforeignlanguage{en}{Human3.{6M}: {Large} {Scale} {Datasets} and
  {Predictive} {Methods} for {3D} {Human} {Sensing} in {Natural}
  {Environments}},'' \emph{\BIBforeignlanguage{en}{PAMI}}, vol.~36, no.~7, pp.
  1325--1339, Jul. 2014. [Online]. Available:
  \url{http://ieeexplore.ieee.org/document/6682899/}
\BIBentrySTDinterwordspacing

\bibitem{Cui_2020_CVPR}
Q.~Cui, H.~Sun, and F.~Yang, ``Learning dynamic relationships for 3{D} human
  motion prediction,'' in \emph{CVPR}, 2020.

\bibitem{KingmaICLR14}
D.~P. Kingma and J.~Ba, ``Adam: {A} method for stochastic optimization,'' in
  \emph{ICLR}, 2015.

\bibitem{vandermaaten08a}
\BIBentryALTinterwordspacing
L.~van~der Maaten and G.~Hinton, ``Visualizing data using t-{SNE},''
  \emph{JMLR}, vol.~9, no.~86, pp. 2579--2605, 2008. [Online]. Available:
  \url{http://jmlr.org/papers/v9/vandermaaten08a.html}
\BIBentrySTDinterwordspacing

\end{thebibliography}
\end{document}